\documentclass[dvipsnames,format=sigconf,authorversion=true]{acmart}

\usepackage{xcolor}
\usepackage{adjustbox}
\usepackage{colortbl}
\usepackage{tabularx}
\usepackage{booktabs} 
\usepackage{graphicx}
\usepackage{subfig}
\usepackage[linesnumbered, vlined, ruled]{algorithm2e}
\SetArgSty{textup}
\usepackage{enumerate}
\usepackage{prettyref}
\usepackage{color, colortbl}
\usepackage{braket}
\definecolor{MyHiLiRow}{gray}{0.9}
\usepackage{tcolorbox}
\usepackage{caption}
\usepackage{seqsplit}
\usepackage{bm}

\usepackage{nameref}
\usepackage{xr}
\usepackage[whole]{bxcjkjatype}

\newcommand{\argmin}{\mathop{\rm argmin}\limits}

\newcommand\tabblue[1]{{\color[HTML]{1f77b4}{#1}}}
\newcommand\taborange[1]{{\color[HTML]{ff7f0e}{#1}}}
\newcommand\tabgreen[1]{{\color[HTML]{2ca02c}{#1}}}
\newcommand\tabred[1]{{\color[HTML]{d62728}{#1}}}
\newcommand\tabpurple[1]{{\color[HTML]{9467bd}{#1}}}
\newcommand\tabbrown[1]{{\color[HTML]{8c564b}{#1}}}
\newcommand\tabpink[1]{{\color[HTML]{e377c2}{#1}}}

\newcommand{\pref}{\prettyref}

\newrefformat{fig}{Figure~\ref{#1}}
\newrefformat{supfig}{Figure~\ref{#1}}
\newrefformat{tab}{Table~\ref{#1}}
\newrefformat{suptab}{Figure~\ref{#1}}
\newrefformat{sec}{Section~\ref{#1}}
\newrefformat{subsec}{Section~\ref{#1}}
\newrefformat{app}{Appendix~\ref{#1}}
\newrefformat{alg}{Algorithm~\ref{#1}}
\newrefformat{property}{Property~\ref{#1}}
\newrefformat{theorem}{Theorem~\ref{#1}}
\newrefformat{corollary}{Corollary~\ref{#1}}
\newrefformat{proposition}{Proposition~\ref{#1}}
\newrefformat{def}{Definition~\ref{#1}}
\newrefformat{eq}{equation~(\ref{#1})}

\newcommand{\myemph}[1]{\textit{#1}}

\AtBeginDocument{%
  \providecommand\BibTeX{{%
    \normalfont B\kern-0.5em{\scshape i\kern-0.25em b}\kern-0.8em\TeX}}}

\setcopyright{acmcopyright}

\copyrightyear{2025}
\acmYear{2025}
\setcopyright{acmlicensed}\acmConference[GECCO '25]{Genetic and Evolutionary Computation Conference}{July 14--18, 2025}{M\'{a}laga, Spain}
\acmBooktitle{Genetic and Evolutionary Computation Conference (GECCO '25), July 14--18, 2025, M\'{a}laga, Spain}
\acmDOI{10.1145/3712256.3726381}
\acmISBN{979-8-4007-1465-8/2025/07}

\begin{document}

\title[Analyzing the Landscape of the ISSP]{Analyzing the Landscape of the Indicator-based Subset Selection Problem}





\author{Keisuke Korogi}
\affiliation{%
  \institution{Yokohama National University}
  \city{Yokohama}
  \state{Kanagawa}
  \country{Japan}}
  \email{keisuke.korogi.52@gmail.com}
  
\author{Ryoji Tanabe}
\affiliation{%
  \institution{Yokohama National University}
  \city{Yokohama}
  \state{Kanagawa}
  \country{Japan}}
  \email{rt.ryoji.tanabe@gmail.com}







\begin{abstract}


The indicator-based subset selection problem (ISSP) involves finding a point subset that minimizes or maximizes a quality indicator.
The ISSP is frequently found in evolutionary multi-objective optimization (EMO).
An in-depth understanding of the landscape of the ISSP could be helpful in developing efficient subset selection methods and explaining their performance.
However, the landscape of the ISSP is poorly understood.
To address this issue, this paper analyzes the landscape of the ISSP by using various traditional landscape analysis measures and exact local optima networks (LONs).
This paper mainly investigates how the landscape of the ISSP is influenced by the choice of a quality indicator and the shape of the Pareto front.
Our findings provide insightful information about the ISSP.
For example, high neutrality and many local optima are observed in the results for ISSP instances with the additive $\epsilon$-indicator. 





\end{abstract}





\maketitle

\section{Introduction}
\label{sec:introduction}

\noindent \textit{General context.}
This paper considers the minimization of $d$ objective functions $f_1, \dots, f_d$.
An evolutionary multi-objective optimization (EMO) algorithm~\cite{Deb01} aims to find a non-dominated solution set that approximates the Pareto front (PF) in the objective space.
Thereafter, this non-dominated solution set is used for an a posteriori decision making~\cite{PurshouseDMMW14}.
In this approach, the decision maker selects a solution from the non-dominated solution set according to her/his preference.
For simplicity, this paper denotes a $d$-dimensional objective vector $f(x)$ of a solution $x$ as a point $p \in \mathbb{R}^d$, i.e., $p=f(x)$.
The remainder of this paper considers only the objective space $V \subseteq \mathbb{R}^d$.



Quality indicators~\cite{ZitzlerTLFF03,LiY19} evaluate the quality of a point set found by EMO algorithms in terms of how well it approximates the PF.
Quality indicators play a crucial role in benchmarking EMO algorithms.
This paper considers the following seven quality indicators: hypervolume (HV) \cite{ZitzlerT98}, inverted generational distance (IGD) \cite{CoelloS04}, IGD plus (IGD$^+$) \cite{Ishibuchi15}, the additive $\epsilon$-indicator ($\epsilon$) \cite{ZitzlerTLFF03}, R2 \cite{HansenJ98}, new R2 (NR2) \cite{Shang18}, and $s$-energy (SE) \cite{HardinS04}.
Generally, each quality indicator prefers a specific distribution of points~\cite{TanabeI20b,LiefoogheD16,JiangOZF14}.


Given a quality indicator $\mathcal{I}$, a non-dominated point set $P \subseteq V$ of size $n$, the indicator-based subset selection problem (ISSP)~\cite{BasseurDGL16} involves finding an optimal subset $S^* \subset P$  of size $k$ that minimizes $\mathcal{I}$, where $k<n$.
This paper denotes an ISSP instance with $\mathcal{I}$ as  $\mathcal{I}$-SSP.
%
The HV-SSP, an ISSP instance with HV, has been well studied in the EMO community~\cite{BaderZ11,GuerreiroFP16,NanSIH23}.
The ISSP with various quality indicators has also been addressed in the literature, e.g., the $\epsilon$-SSP~\cite{BringmannFK14ppsn,BringmannFK15}, IGD-SSP~\cite{ChenIS22,NanISS24}, IGD$^+$-SSP \cite{ChenIS22,NanISS24}, and NR2-SSP~\cite{ShangIC21}.
In EMO, the ISSP appears in environmental selection in indicator-based EMO algorithms~\cite{Falcon-CardonaC20} and the postprocessing of the unbounded external archive~\cite{BringmannFK14ppsn,TanabeIO17} that maintains all non-dominated points found so far.
In the former case, $P$ in the ISSP is the union of the population and offspring.
In the latter case, $P$ is the unbounded external archive, which could include a large number of non-dominated points.

Since the ISSP is an NP-hard problem~\cite{BringmannCE17}, inexact approaches are effective and practical for the ISSP.\footnote{Only when $d=2$, dynamic programming can find an optimal subset on the HV-SSP and $\epsilon$-SSP in a reasonable computation time~\cite{BringmannFK14}.}
Representative inexact methods for the ISSP include greedy search~\cite{BasseurDGL16,GuerreiroFP16,ShangIC21,ChenIS22} and local search~\cite{BasseurDGL16,BradstreetBW06,NanSIH23}.
In \cite{BasseurDGL16}, Basseur et al. presented three greedy search methods and a local search method for the HV-SSP, but they can be straightforwardly applied to the ISSP with any quality indicator.
Some advanced greedy search and local search methods for the ISSP have been proposed in recent studies (e.g., \cite{ShangIC21,ChenIS22,NanSIH23}).
These methods exploit the explicit property of each quality indicator (e.g., the submodular property of HV, IGD, and IGD$^+$).
Thus, unlike the four methods in \cite{BasseurDGL16}, such advanced methods can be applied to only the ISSP with a particular quality indicator.


\vspace{0.2em}
\noindent \textit{Motivation.}
Landscape analysis provides a better understanding of the property of a problem~\cite{PitzerA12,RichterE14,MalanE13}.
Landscape analysis methods identify landscape features, including the number of global and local optima, distribution of local optima, neutrality, and basins of attraction.
An in-depth understanding of the landscape of a problem could be useful for developing efficient optimization methods and predicting the performance of methods~\cite{OchoaVDT14}.
%
In addition, it is important to understand how the landscape of a problem instance is characterized by parameters for the instantiation.
For example, two previous studies~\cite{LiefoogheTV24,DanchezDiazMM24} investigated how the landscape of the feature selection problem is influenced by the existence of the regularization and the type of machine learning model.
Their findings indicate the type of difficulty produced by each problem's element.
This is also important for benchmarking feature selection methods.





Unfortunately, very little is known about the landscape of the ISSP.
Multi-objective landscape analysis is a hot research topic in the EMO community~(e.g., \cite{SchapermeierGK22,LiefoogheOVD23,RodriguezTAB24,OchoaLV24}).
Landscapes of subset selection problems (e.g., the feature selection problem~\cite{MostertMOE19,LiefoogheTV24,DanchezDiazMM24}) have also been analyzed.
However, landscape analysis of the ISSP has not received any attention.

\vspace{0.2em}
\noindent \textit{Contributions.}
Motivated by the above discussion, this paper analyzes the landscape of the ISSP by means of various traditional landscape measures (e.g., fitness distance correlation~\cite{JonesF95} and counting the number of local optima) and local optima networks (LONs)~\cite{OchoaTVD08,OchoaVDT14}.
This paper mainly aims to understand the influence of the type of quality indicator and shape of the PF on the landscape of the ISSP.
For this purpose, we consider the above-mentioned seven quality indicators (HV, IGD, IGD$^+$, $\epsilon$, R2, NR2, and SE) and seven PFs. 
We also investigate the scalability of landscape features with respect to the number of objectives $d$ as well as the relation between landscape features and the performance of subset selection methods.




\vspace{0.2em}
\noindent \textit{Contributions.}
Section \ref{sec:preliminaries} provides the necessary preliminaries.
Section \ref{sec:setting} describes the experimental setup.
Section \ref{sec:results} presents the analysis results. 
Section \ref{sec:conclusion} concludes the paper.

\section{Preliminaries}
\label{sec:preliminaries}

\subsection{Multi-objective optimization}
\label{subsec:prelim:mo}

Multi-objective optimization aims to simultaneously minimize $d$ objective functions $f = (f_1, \dots, f_d)$.
Let $V \subseteq \mathbb{R}^d$ denote the $d$-dimensional objective space.
As described in Section \ref{sec:introduction}, an objective vector $f(x)$ in $V$ is denoted as a point $p$ in this paper, i.e., $p=f(x)$, and $p=(p_1, \dots, p_d)^{\top}$.


Considering two points $p$ and $q \in V$, $p$ is said to \myemph{dominate} $q$ if $p_i \leq q_i$ for all $i \in \{1, \dots, d\}$ and $p_i < q_i$ for at least one index $i$.
This Pareto dominance relation between $p$ and $q$ is denoted by $p \prec q$.
Similarly, $p$ is said to \myemph{weakly dominate} $q$, denoted by $p \preceq q$ if $p_i \leq q_i$ for all $i \in \{1, \dots, d\}$.
In addition, $p^* \in V$ is a \myemph{Pareto optimal point} if $p^*$ is not dominated by any point in $V$.
The \myemph{Pareto front (PF)} is the set of all Pareto optimal points $\{ p^* \in V \mid {}^\nexists p \in V,\, p \prec p^* \}$.

\subsection{Quality indicators}
\label{subsec:prelim:qi}

Let $P \subset V$ be a set of $n$ non-dominated points found by an EMO algorithm.
It is desirable that $P$ approximates the PF well.
Let also $\Omega$ be the family of all non-dominated point sets in $V$.
A \myemph{quality indicator} $\mathcal{I} \colon \Omega \to \mathbb{R}$  evaluates the quality of $P$ in terms of at least one of the following three aspects: convergence, uniformity, and spread~\cite{ZitzlerTLFF03,LiY19}.
This paper focuses only on the unary quality indicator that maps a single point set to a scalar value.
The convergence of a point set $P$ means the closeness of points in $P$ to the PF.
The uniformity of $P$ represents how even the distribution of points in $P$ is.
$P$ has a good spread if points in $P$ cover the PF well.
The combination of uniformity and spread is generally called diversity in the EMO literature.
Since we focus only on the comparison of non-dominated point sets of the same size $n$, we do not consider the cardinality of $P$.
%
A quality indicator $\mathcal{I}$ is Pareto-compliant if the ranking of all point sets in $\Omega$ by $\mathcal{I}$ is consistent with the Pareto dominance relation~\cite{KnowlesTZ06}.


Below, we briefly describe the following seven quality indicators mentioned in \pref{sec:introduction}: HV~\cite{ZitzlerT98}, IGD~\cite{CoelloS04}, IGD$^+$~\cite{Ishibuchi15}, $\epsilon$~\cite{ZitzlerTLFF03}, R2~\cite{HansenJ98}, NR2~\cite{Shang18}, and SE~\cite{HardinS04}.
All the quality indicators except SE evaluate the quality of point sets in terms of both convergence and diversity, whereas SE considers only the diversity of point sets.
While HV and NR2 are to be maximized, the others are to be minimized.
Although HV is Pareto-compliant, HV prefers a set of non-uniformly distributed points in some cases~\cite{TanabeI20b,AugerBBZ09,JiangOZF14}.
HV also requires high computational cost when $d$ is large.
For this reason, other quality indicators are used to complement HV.


HV measures the volume of the union of regions that are dominated by $P$ and bounded by the reference point $r \in V$.
IGD, IGD$^+$, and $\epsilon$ use the reference point set $R$, where the points in $R$ are uniformly distributed on the PF.
IGD calculates the distance from each point in $R$ to its nearest point in $P$.
IGD$^+$ is a weakly Pareto-compliant version of IGD.
The only difference between IGD and IGD$^+$ is the type of distance function.
The unary version of $\epsilon$ measures the minimum shift such that each point in $P$ weakly dominates at least one reference point in $R$.
R2 and NR2 require a weight vector set $W$.
R2 calculates the average of the minimum weighted Tchebycheff function values of $P$.
NR2 is an improved version of R2 for better HV approximation, where NR2 requires the reference point $r$ as in HV.
SE measures the sum of the reciprocal of the distance between all pairs of points in $P$.

\subsection{Indicator-based subset selection problem}
\label{subsec:prelim:issp}

Given a non-dominated point set $P$ of size $n$ and a quality indicator $\mathcal{I}$, the \myemph{ISSP}  involves finding a subset $S^* \subset P$ of size $k$ with the minimum quality value:
\[
S^* = \argmin_{S \subset P,\, |S| = k} \mathcal{I}(S),
\]
where $k < n$.
The number of all feasible subsets is $\binom{n}{k}$. 
When $\mathcal{I}$ is to be maximized (e.g., HV and NR2), $\mathcal{I}$ is reformulated as $-\mathcal{I}$.

In general, $S$ in the ISSP is represented by an $n$-dimensional binary vector $x = (x_1, \dots, x_n)^\top \in \{0,1\}^n$, where $\sum^n_{i=1} x_i = k$.
For each $i \in \{1, \ldots, n\}$, the $i$-th point $p_i$ in $P$ is included in the subset $S$ if $x_i=1$.
For example, consider $n=4$, $k=2$, and $x = (1, 0, 1, 0)^\top$.
In this case, $S = \{p_1, p_3\}$. 
While a subset $S$ is a phenotype, a binary vector $x$ is a genotype.
For simplicity, the reminder of this paper denotes $x$ as a \myemph{solution} of the ISSP.

\subsection{Landscape analysis}
\label{subsec:prelim:land-analysis}

We define the \myemph{fitness landscape} of the ISSP by a 3-tuple as follows: $(\mathcal{X}, \mathcal{N}, \mathcal{I})$.
Here, $\mathcal{X}$ is the solution space of the ISSP, \textit{not} that of the multi-objective optimization problem.
The size of $\mathcal{X}$ is equivalent to the number of all feasible solutions, i.e., $|\mathcal{X}|=\binom{n}{k}$.
$\mathcal{N} \colon \mathcal{X} \to 2^{\mathcal{X}}$ is the \myemph{neighborhood relation} defined by the minimum possible movement.
We use the 2-bit-swap neighborhood relation such that the size of a subset is always $k$.
Let us consider two binary vectors $x_1$ and $x_2 \in \{0,1\}^n$, where each of them represents a subset in the ISSP.
We say that $x_1$ is in the neighborhood of $x_2$ if $x_1$ becomes identical to $x_2$ after selecting two bits of $x_1$, one with value 0 and one with value 1, and swapping their values.
In other words, $x_1$ is said to be in the neighborhood of $x_2$ if the Hamming distance between $x_1$ and $x_2$ is two. 
For example, for $n=4$ and $k=2$, when $x_1 = (1, 0, 1, 0)^{\top}$ and $x_2 = (1, 0, 0, 1)^{\top}$, $x_1$ is identical with $x_2$ after swapping the third and fourth bits in $x_1$.
Therefore, $x_1$ is in $\mathcal{N}(x_2)$.
The size of the neighborhood $|\mathcal{N}(x)|$ of any solution $x$ is $k (n-k)$, which is the number of ways to select bits with value 0 and 1. 
In the ISSP, a quality indicator $\mathcal{I}$ evaluates a subset $S$, which is represented by a solution $x$.
Thus, $\mathcal{I}$ can be interpreted as a fitness function for $x$, i.e., $\mathcal{I}(x)$ is the fitness value of $x$.



A solution $x^* \in \mathcal{X}$ is a \myemph{global optimum} if 
$\mathcal{I}(x^*) \leq \mathcal{I}(x)$ for any $x \in \mathcal{X}$.
There could be more than one global optima.
A solution $x^{\mathrm{loc}} \in \mathcal{X}$ is a \myemph{local optimum} if $\mathcal{I}(x^{\mathrm{loc}}) \leq \mathcal{I}(x)$ for any $x \in \mathcal{N}(x^{\mathrm{loc}})$.
The distribution of local optima plays a central role in determining the structure of the landscape.


Starting from an initial solution $x^{\mathrm{init}}$, \myemph{local search} explores the neighborhood $\mathcal{N}(x)$ by the 2-bit-swap operation and moves to a better solution until no solution in $\mathcal{N}(x)$ can improve $\mathcal{I}$.
Thus, local search can be considered as a function that maps $x^{\mathrm{init}}$ to a local optimum $x^{\mathrm{loc}}$, i.e., $x^{\mathrm{loc}}=\mathtt{local\_search}(x^{\mathrm{init}})$.
The \myemph{basin of attraction} $\mathcal{B}$ \cite{OchoaTVD08} of a local optimum $x^{\mathrm{loc}}$ is the set of solutions that converge to $x^{\mathrm{loc}}$ by performing local search, i.e., $\mathcal{B}(x^{\mathrm{loc}}) = \{x \in \mathcal{X} \: | \: x^{\mathrm{loc}}=\mathtt{local\_search}(x)\}$.

A \myemph{plateau} is a part of the solution space where the quality indicator value is not changed.
In other words, no search direction is available on plateaus.
Formally, a plateau is a maximal subset $\mathcal{X}^{\mathrm{pla}} \subseteq \mathcal{X}$ that satisfies with the following conditions:
for any two distinct solution $x,\, x' \in \mathcal{X}$, there exists a sequence of $m$ solutions $(x=) x_1, ..., x_m (=x')$ such that $x_{i+1} \in \mathcal{N}(x_i)$ and $\mathcal{I}(x_i) = \mathcal{I}(x_{i+1})$ for $i = 1, ..., m-1$. 
%
\myemph{Global and local optima plateaus} are plateaus that include global and local optima, respectively.

A \myemph{neutrality} is defined as the average proportion of the number of solutions with the same quality in the neighborhood of each solution as follows:
\[
\frac{1}{|\mathcal{X}|} \sum_{x \in \mathcal{X}} \frac{\bigl| \{ x' \in \mathcal{N}(x) \mid \mathcal{I}(x) = \mathcal{I}(x') \} \bigr| }{|\mathcal{N}(x)|}.
\]

This paper defines the \myemph{ruggedness} in the landscape as the Spearman's rank correlation coefficient between $\mathcal{I}(x)$ and $\mathcal{I}(x')$, where $x \in \mathcal{X}$, $x' \in \mathcal{N}(x)$.
A large correlation coefficient means that the landscape is smooth.

\myemph{Fitness distance correlation (FDC)}~\cite{JonesF95} measures Spearman's rank correlation coefficient between the distance of solutions to the nearest global optimum and quality indicator values.
FDC measures the global structure of the landscape.
The Hamming distance is generally used to measure the distance between two binary solutions $x^1$ and $x^2$: $\mathtt{dist}_{H}(x^1, x^2) = |\{ i \in \{1, \dots, n\} \mid x^1_i \neq x^2_i \}|$.
In addition, we use the (1-)Wasserstein distance (also known as the earth mover's distance) to measure the distance between two subsets $S^1 = \{ p^1_1, \dots, p^1_k \}$ and $S^2 = \{p^2_1, \dots, p^2_k\}$:
\[
\mathtt{dist}_{W}(S^1, S^2) = \min_{\sigma} \sum_{i \in \{1, \dots, k\}} \left\| p^1_{i} - p^2_{\sigma(i)} \right\|_2,
\]
where $\sigma$ runs over all permutations of size $k$. 
The Wasserstein distance is useful for measuring the similarity between two sets and is used in recent studies for machine learning \cite{ArjovskyCB17,FrognerZMAP15}.
While the Hamming distance is based on the genotype space (i.e., the binary search space), the Wasserstein distance is based on the phenotype space (i.e., the objective space $V$).
In the ISSP, the number of possible Hamming distance values is at most $k+1$.
However, this is not the case for the Wasserstein distance.


%

%

\myemph{LONs}~\cite{OchoaTVD08,OchoaVDT14} characterize the landscape of a problem by a complex network.
LONs mainly focus on the distribution of local optima. 
A LON is defined by a weighted directed graph $\mathcal{G} = (\mathcal{V}, \mathcal{E})$, where $\mathcal{V}$ is a set of nodes, and $\mathcal{E}$ is a set of edges.
Each node $v$ in $\mathcal{V}$ represents a local optimum $x^{\mathrm{loc}}$.
The width of each node $v$ represents the size of the basin of attraction $|\mathcal{B}(v)|$.

An \myemph{escape edge} $e$~\cite{VerelDOT11} is defined by the distance function $\mathtt{dist}$ and a positive integer $D > 0$.
We use the Hamming distance as $\mathtt{dist}$.
In this case, the minimum $D$ value is $4$.
%
Let us consider two nodes $v_i$ and $v_j$, which correspond to the $i$-th and $j$-th local optima $x^{\mathrm{loc}}_i$ and $x^{\mathrm{loc}}_j$, respectively.
An escape edge $e_{i, j}$ exists between two nodes $v_i$ and $v_j$ if there exists a solution $x \in \mathcal{X}$ such that $\mathtt{dist}(x, x^{\mathrm{loc}}_i) \leq D$ and $\mathtt{local\_search}(x) = x^{\mathrm{loc}}_j$.
The weight $w_{i, j}$ of $e_{i, j}$ is the cardinality of a set as follows: $w_{i, j} = |\{ x \in \mathcal{X} \mid \mathtt{dist}(x, x^{\mathrm{loc}}_i) \leq D \ \mathrm{and}\  \mathtt{local\_search}(x) = x^{\mathrm{loc}}_j \}|$. 
This weight $w_{i, j}$ is generally normalized by  $|\{ x \in \mathcal{X} \mid \mathtt{dist}(x, x^{\mathrm{loc}}_i) \leq D \}|$.





\section{Experimental setup}
\label{sec:setting}




Unless otherwise noted, the number of objectives $d$, point set size $n$, subset size $k$ were fixed as follows: $d=3$, $n = 50$, and $k = 5$.
In this setting of $n$ and $k$, the number of all subsets is $\binom{n}{k} = 2\,118\,760$, which allows us to fully enumerate all subsets and compute landscape measures.
Note that the enumeration of all solutions is needed to construct exact LONs.
Landscape analysis of the ISSP with larger numbers of $n$ and $k$ is an avenue for future work.



As in \cite{NanSIH23}, this paper focuses on the following six PFs: a linear PF (DTLZ1~\cite{DebTLZ05}), concave PF (DTLZ2), convex PF (convDTLZ2~\cite{DebJ14}), and their inverted versions (inv-linear, inv-nonconvex, and inv-convex PFs).
In addition, this paper considers a discontinuous PF (DTLZ7).
All seven PFs are normalized into $[0,1]^d$.
A non-dominated point set $P$ of size $n$ was generated on each PF as follows.
First, all $n$ points in $P$ were uniformly generated on the linear PF by the method proposed in \cite{BlankDDBS21}, which is implemented in \texttt{pymoo}~\cite{BlankD20}.
For each $p \in P$, $\sum_{i=1}^{d} p_i = 1$ on the linear PF.
Then, except for the linear and discontinuous PFs, $P$ is translated for each PF using the method presented by the method described in \cite{TianXZCJ18}.
$P$ for the discontinuous PF was generated by the special method proposed in \cite{TianXZCJ18}.
Due to the combinatorial property of this generation method, $n$ was set to $49$ only for the discontinuous PF.

We used the following seven quality indicators described in Section \ref{subsec:prelim:qi}: HV, IGD, IGD$^+$, R2, NR2, $\epsilon$, and SE.
Since $P \subset [0, 1]^d$, the reference point $r \in \mathbb{R}^d$ for HV and NR2 was set to $r = (1.1, \dots, 1.1)^\top$.
For each PF, $P$ was used as the reference point set $R$ for IGD, IGD$^+$, and $\epsilon$.
We also used a set of $n$ uniformly distributed weight vectors $W$ for R2 and NR2.


\section{Results}
\label{sec:results}

As in \cite{LiefoogheTV24}, our analysis is based on various traditional landscape measures and LONs.
First, \pref{subsec:distr-qi} investigates the distribution of quality indicator values on the 49 ISSP instances, where this experiment uses the 7 quality indicators and the 7 PFs.
Then, \pref{subsec:corr-sol-rank} analyzes the correlation of subset rankings on the 49 ISSP instances.
Sections \ref{subsec:num-optima}, \ref{subsec:ruggedness}, and \ref{subsec:neutrality} focus on the numbers of global and local optima, ruggedness, and neutrality, respectively.
Sections \ref{subsec:fdc} and \ref{subsec:lons} analyze the landscape of the ISSP by using FDC and LONs, respectively. 
\pref{subsec:alg} investigates the relation between the performance of subset selection methods and the landscape of the ISSP.
Finally, \pref{subsec:d} analyzes the influence of $d$ on the landscape of the ISSP.

\noindent \textit{Meaning of color and order in Figures \ref{fig:distr-qi}, \ref{fig:num-optima}, \ref{fig:rug}, \ref{fig:neu}, \ref{fig:fdc}, and \ref{fig:alg}.}
For each ISSP instance, the seven box plots in Figures \ref{fig:distr-qi} and \ref{fig:alg} show the results for the \tabblue{$\blacksquare$} linear, \taborange{$\blacksquare$} convex, \tabgreen{$\blacksquare$} nonconvex, \tabred{$\blacksquare$} inv-linear, \tabpurple{$\blacksquare$} inv-convex, \tabbrown{$\blacksquare$} inv-nonconvex, and \tabpink{$\blacksquare$} discontinuous PFs from left to right in the figures, respectively.
The same is true for the seven bar plots in Figures \ref{fig:num-optima}--\ref{fig:fdc}.

\subsection{Distribution of quality indicator values}
\label{subsec:distr-qi}

\begin{figure}[t]
    \centering
    \includegraphics[width=.93\linewidth]{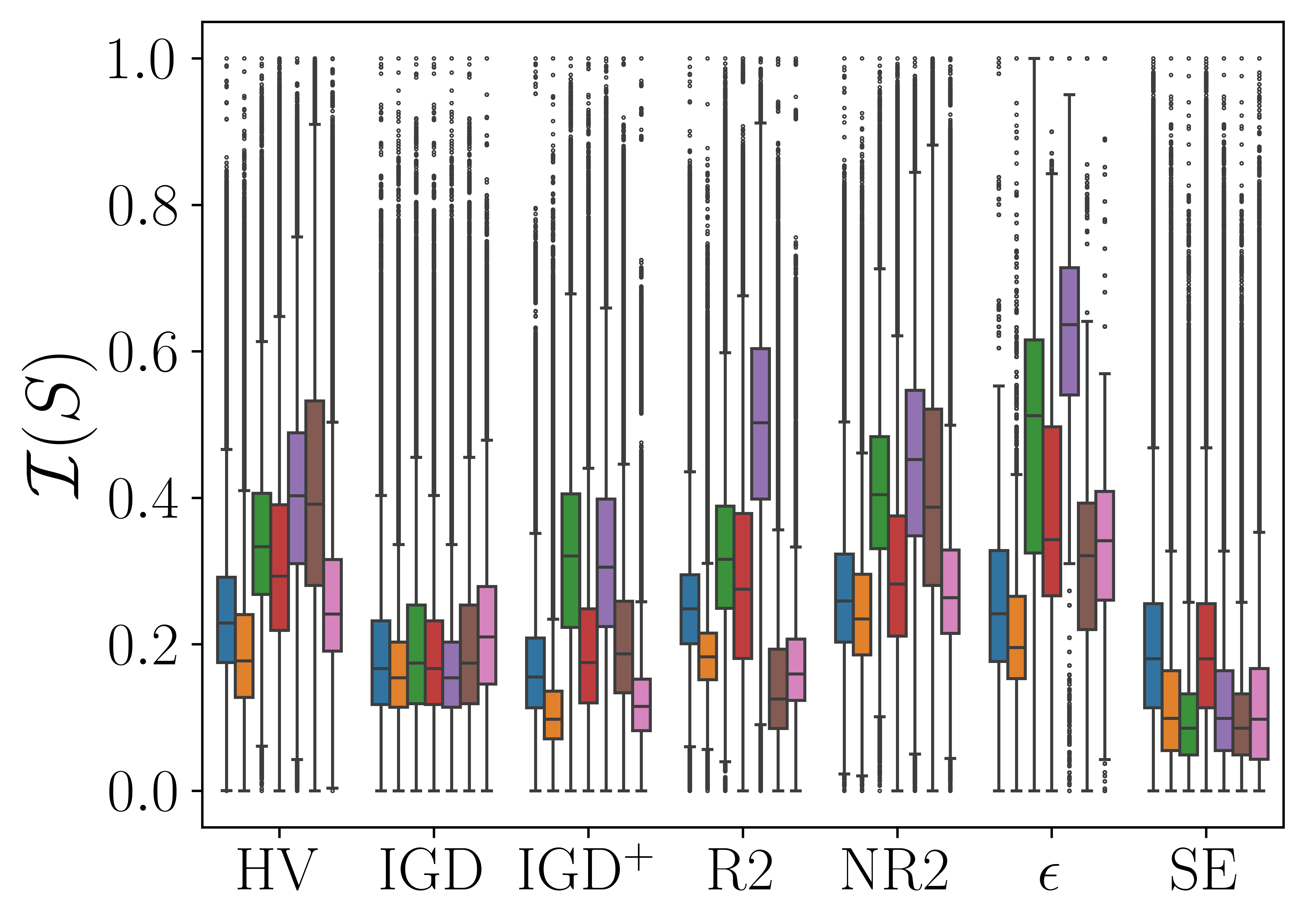}
    \caption{Distribution of quality indicator values $\mathcal{I}(S)$, where those are min-max normalized.} 
    \label{fig:distr-qi}
\end{figure}

\pref{fig:distr-qi} shows the distribution of normalized quality indicator values of all $2\,118\,760$ solutions (or subsets) on the ISSP with the seven quality indicators and seven PFs.
For readability, we normalized each quality indicator value into $[0,1]$ based on the minimum and maximum values.
Recall that all quality indicators are to be minimized in this work.


As seen from \pref{fig:distr-qi}, the distribution of quality indicator values depends on the type of quality indicator and PF.
The results suggest that the solution space of the IGD-SSP and SE-SSP includes many high-quality subsets.
For the IGD-SSP and SE-SSP, the distributions are the same for each of the three conventional PFs (the linear, convex, and non-convex PFs) and their inverted versions (the inv-linear, inv-convex, and inv-nonconvex PFs).
This is because the distance calculation in IGD and SE is invariant with respect to the inversion in the objective space.
%
In the HV-SSP, most quality indicator values for each conventional PF are smaller than those for its corresponding inverted version.
The results also indicate that the solution space of the ISSP includes a relatively large number of poor subsets for a particular combination of quality indicator and PF.
For example, there are many poor subsets and outliers in the distributions of the R2-SSP and $\epsilon$-SSP with the inv-convex PF.


\begin{tcolorbox}[sharpish corners, top=0pt, bottom=0pt, left=2pt, right=2pt, boxrule=0.0pt, colback=black!5!white,leftrule=0.75mm,]
\textbf{Main finding}: The type of quality indicator and PF significantly influences the distribution of quality indicator values in the ISSP.
For example, the quality of a randomly generated subset is likely to be high  for the IGD-SSP and SE-SSP, but this is not true for the R2-SSP and $\epsilon$-SSP with some PFs.



\end{tcolorbox}

\subsection{Correlation of subset rankings}
\label{subsec:corr-sol-rank}

\begin{figure*}[t]
    \centering
    \includegraphics[width=.8\linewidth]{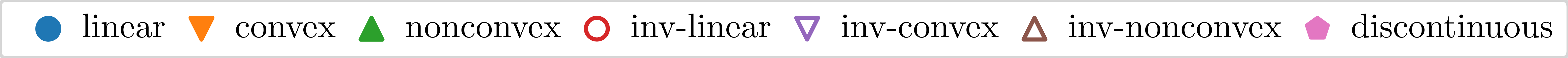}
    \includegraphics[width=.87\linewidth]{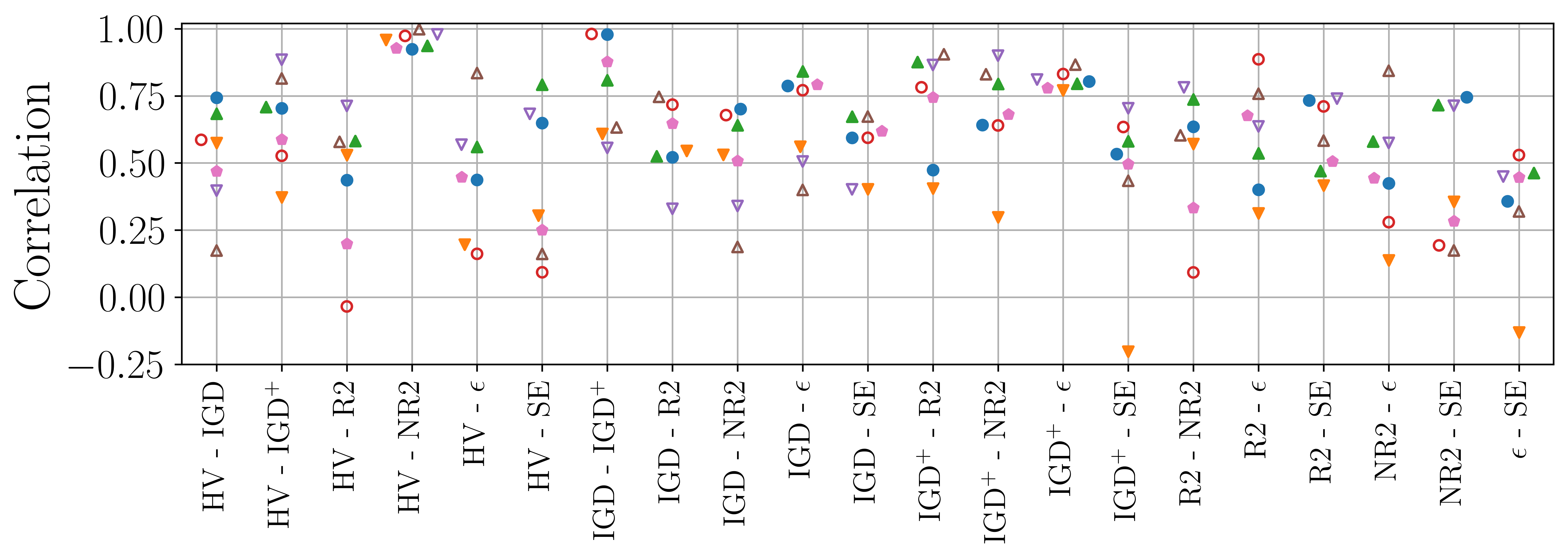}
    \caption{Spearman's rank correlation coefficients between the rankings of all subsets by each pair of quality indicators.}
    \label{fig:corr-sol-rank}
\end{figure*}

\pref{fig:corr-sol-rank} shows Spearman's rank correlation coefficient between the rankings of all subsets by each pair of quality indicators.
For each pair, the seven points represent the results for the seven PFs, respectively.

\begin{figure}[t]
    \centering
    \subfloat[Global optima (plateaus)]{
    \includegraphics[width=.82\linewidth]{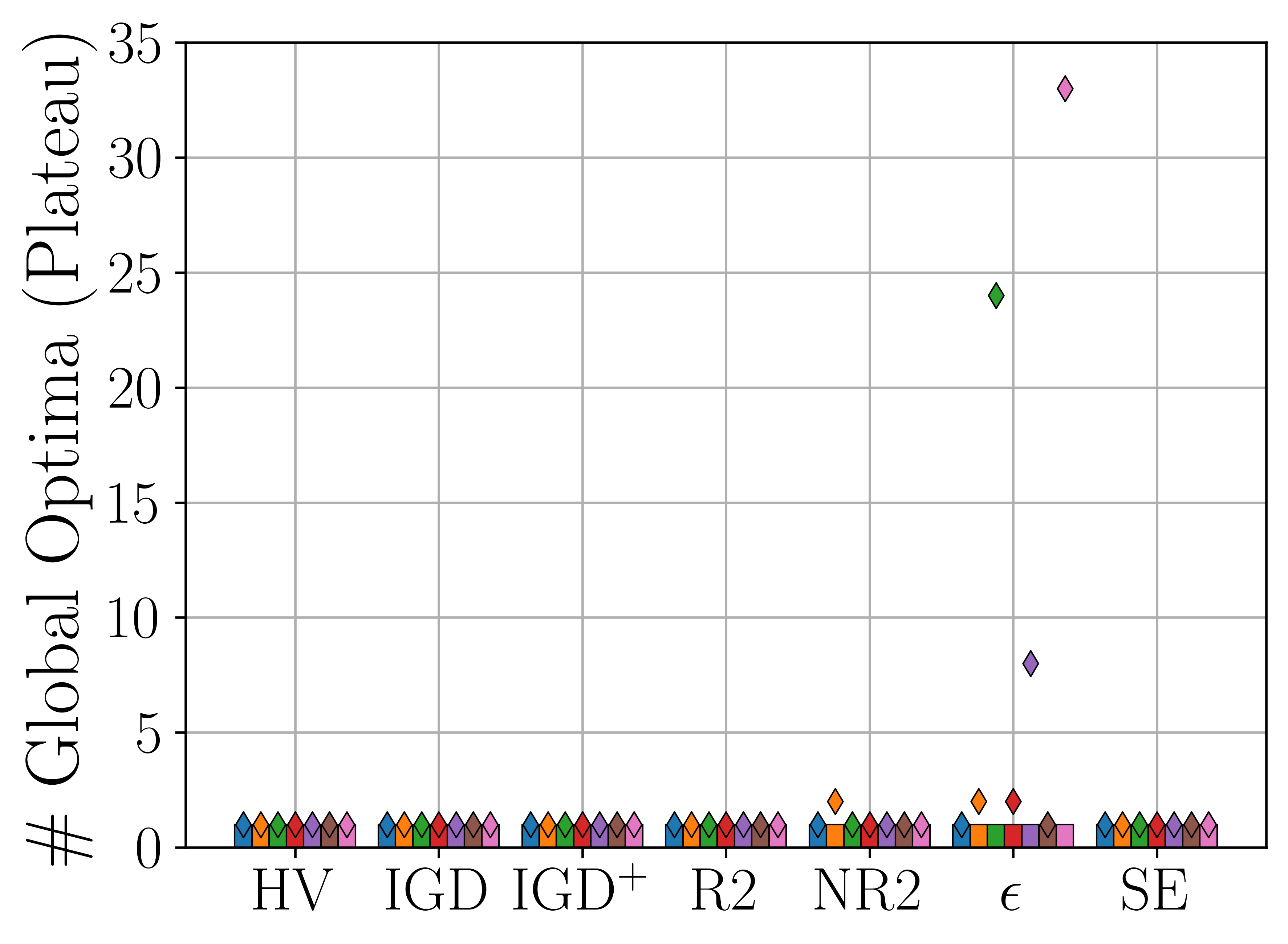}
    }
    \\
    \subfloat[Local optima (plateaus)]{
    \includegraphics[width=.82\linewidth]{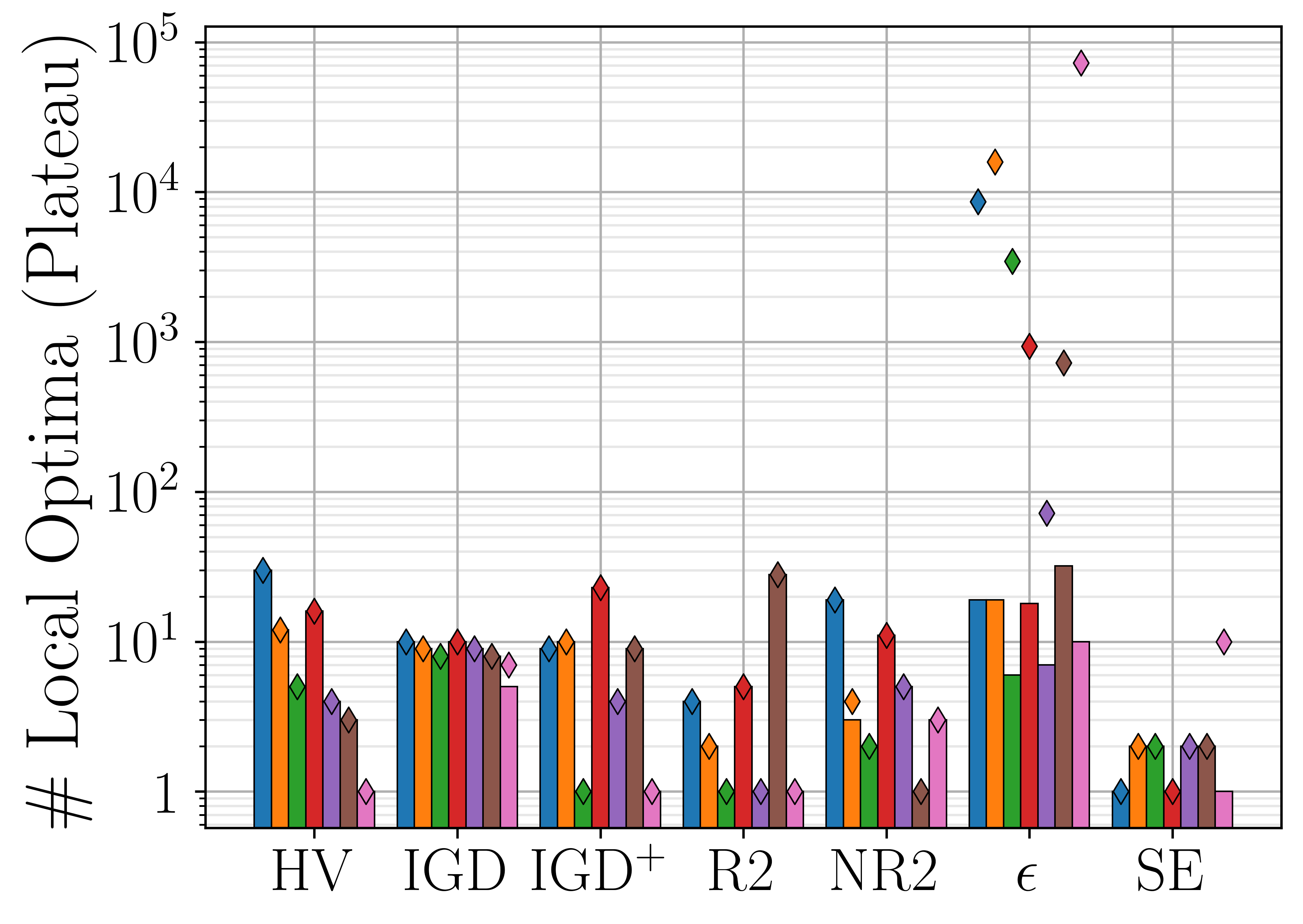}
    }
    \caption{Number of global and local optima (plateaus).} 
    \label{fig:num-optima}
\end{figure}

As shown in \pref{fig:corr-sol-rank}, except for the three cases, the correlation coefficients are positive.
However, the correlation coefficients significantly differ depending on the pair of quality indicators and the shape of the PF.
%

\noindent \textit{Positive correlation.}
As shown in \pref{fig:corr-sol-rank}, 
high correlations are found between the HV-SSP and NR2-SSP independently of the shape of the PF.
This means that a subset $S$ is good (or bad) on the NR2-SSP when $S$ is good (or bad) on the HV-SSP.
This observation proves the rationality of an HV subset selection method proposed in \cite{ShangIC21}, which uses NR2 as a substitute for HV.
While the correlation coefficient of the pair of the IGD-SSP and IGD$^+$-SSP is high for the linear and inv-linear PFs, it is relatively small for the other non-linear PFs.
Our results are partially consistent with \cite{TanabeI20b}, which demonstrated that the optimal $\mu$-distributions of IGD and IGD$^+$ are different for non-linear PFs.
High correlations are also found between the IGD$^+$-SSP and $\epsilon$-SSP.


\noindent \textit{Negative and weak correlation.}
Negative correlations are found in the following three cases: (i) the pair of the HV-SSP and R2-SSP with the inv-linear PF, (ii) the pair of the IGD$^+$-SSP and SE-SSP with the convex PF, and (iii) the pair of the $\epsilon$-SSP and SE-SSP with the convex PF.
The first case is because R2 prefers outer points on the PF, while HV does not necessarily do so in some cases~\cite{TanabeI20b}.
The same is true for the pair of the NR2-SSP and SE-SSP, which shows a weak correlation for the inv-linear and inv-nonconvex PFs. 
The second and third cases are because $n$ points in a point set $P$ generated by the method in \cite{TianXZCJ18} are biasedly distributed to the center of the convex PF, which can be found from the figures in ~\cite{TanabeI20b}.
In addition, IGD$^+$ and $\epsilon$ prefer a part of the objective space where reference points are densely distributed, while SE prefers outer points on the PF.


\begin{tcolorbox}[sharpish corners, top=2pt, bottom=2pt, left=4pt, right=4pt, boxrule=0.0pt, colback=black!5!white,leftrule=0.75mm,]
\textbf{Main finding}: High correlations are observed between specific ISSP instances (e.g., the IGD$^+$-SSP and $\epsilon$-SSP) independently of the shape of the PF.
Since two highly correlated ISSP instances could substitute for each other, it may be effective to reuse a good subset for one as an initial subset for the other.



\end{tcolorbox}

\subsection{Number of global and local optima}
\label{subsec:num-optima}

Figures \ref{fig:num-optima}(a) and (b) show the number of global and local optima on all ISSP instances, respectively.
In \pref{fig:num-optima}, for each ISSP, a point indicates the number of global and local optima, and a bar represents the number of global and local optima plateaus.

\noindent \textit{Number of global optima.}
As shown in \pref{fig:num-optima}(a), all ISSP instances have a single global optima plateau.
The number of global optima is one except for the NR2-SSP and $\epsilon$-SSP, which have multiple global optima depending on the shape of the PF.
Many global optima are observed in the $\epsilon$-SSP, where its numbers are 24, 8, and 33, for the nonconvex, inv-convex, and discontinuous PFs, respectively.

\begin{figure}[t]
    \centering
    \includegraphics[width=.41\linewidth]{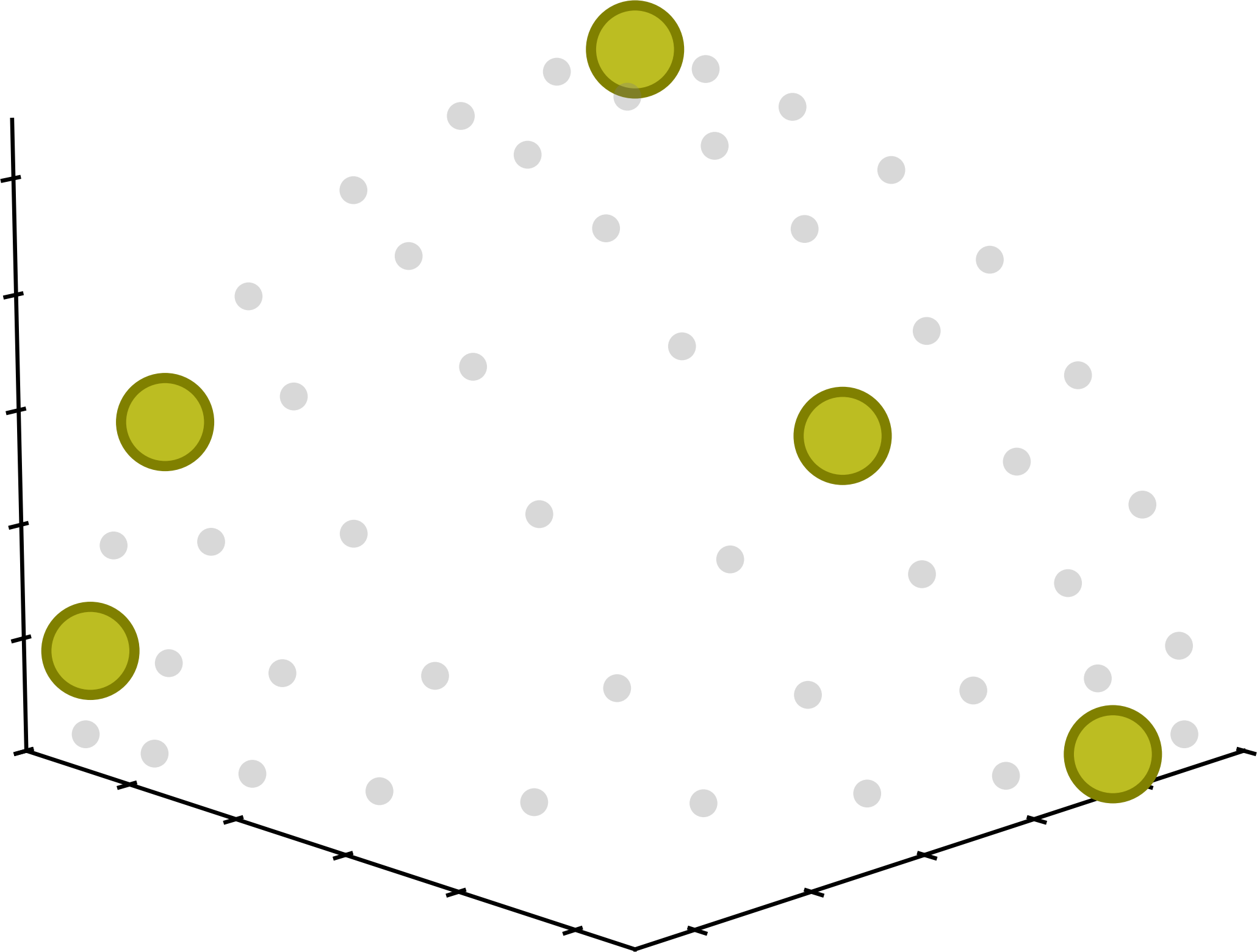}
    \includegraphics[width=.41\linewidth]{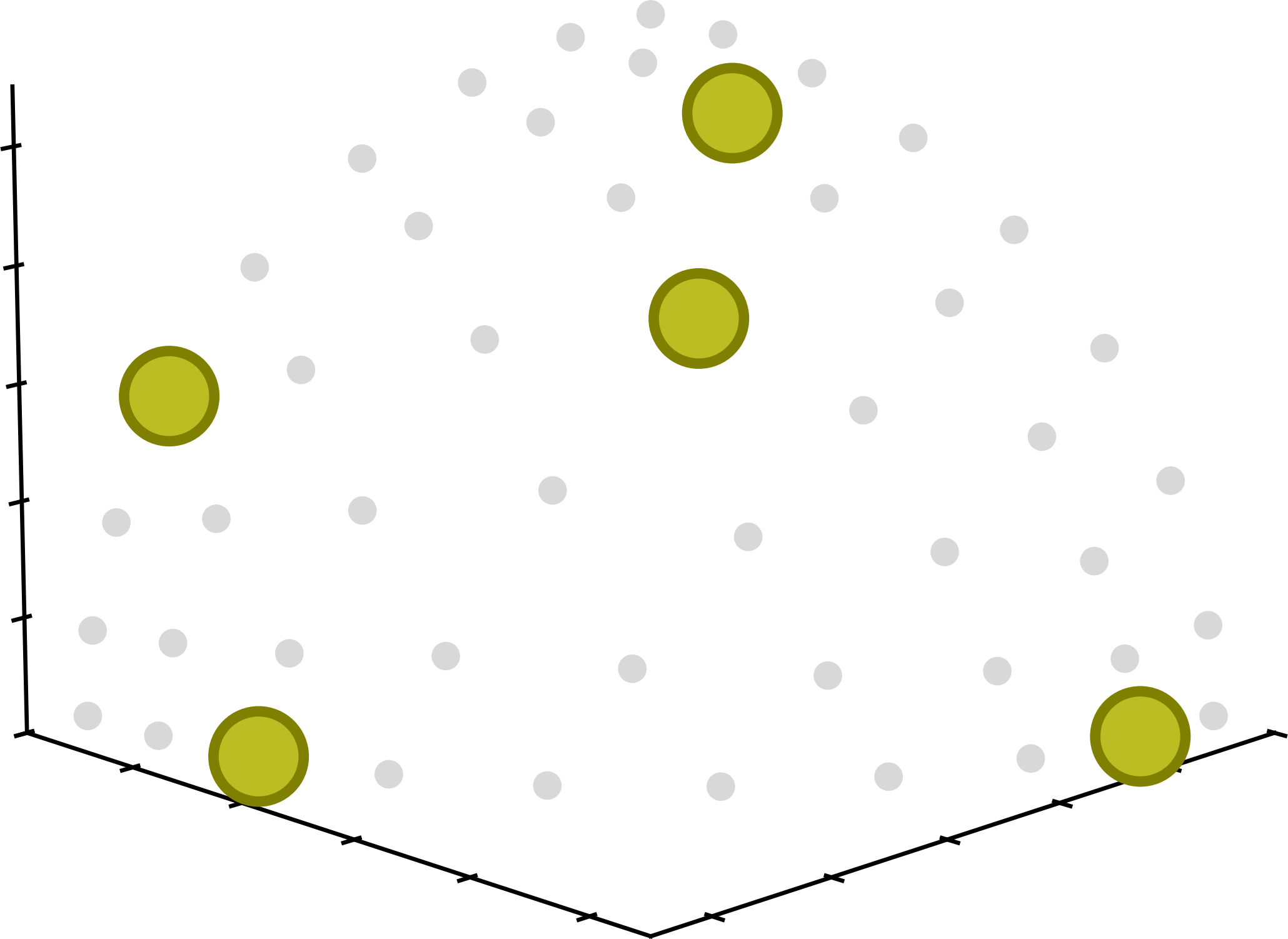}
    \caption{Distributions of two global optimal subsets on the $\epsilon$-SSP with the nonconvex PF. 
    }
    \label{fig:optimal_subset_epsilon}
\end{figure}

\pref{fig:optimal_subset_epsilon} shows the distributions of two global optimal subsets on the $\epsilon$-SSP with the nonconvex PF, where they are on the same global optima plateau.
In \pref{fig:optimal_subset_epsilon}, the yellow large $5$ points are included in a global optimal subset $S^*$, and the gray small $45 (= 50-5)$ points are a set of unselected points $P \setminus S^*$.
As shown in \pref{fig:optimal_subset_epsilon}, the distributions of the two global optimal subsets are totally different.


\noindent \textit{Number of local optima.}
As clearly shown in \pref{fig:num-optima}(b), the $\epsilon$-SSP has a larger number of local optima than the other ISSP instances. 
Notice that the y-axis in \pref{fig:num-optima}(b) is in log scale.
Especially, the results on the $\epsilon$-SSP show that the number of local optima in each conventional PF is larger than that in its inverted version.

Surprisingly, only one local optimum is found on the HV-SSP, IGD$^+$-SSP, R2-SSP, NR2-SSP, and SE-SSP for specific PFs.
This means that their landscapes are unimodal.
Unlike the other ISSP instances, the number of local optima in the IGD-SSP is not significantly influenced by the shape of the PF.

As seen from \pref{fig:num-optima}(b), the numbers of local optima and local optima plateau are identical in most ISSP instances, except for the $\epsilon$-SSP.
This means that the local plateau is seldom found in the ISSP.



\begin{tcolorbox}[sharpish corners, top=2pt, bottom=2pt, left=4pt, right=4pt, boxrule=0.0pt, colback=black!5!white,leftrule=0.75mm,]
\textbf{Main finding}:
The $\epsilon$-SSP is likely to be highly multimodal.
In contrast, the landscape of IGD$^+$-SSP, R2-SSP, NR2-SSP, and SE-SSP could be unimodal for some PFs.
The plateau is unlikely to be a typical landscape feature
in the ISSP, except for the $\epsilon$-SSP.

\end{tcolorbox}

\subsection{Ruggedness}
\label{subsec:ruggedness}


\begin{figure}[t]
    \centering
    \includegraphics[width=.87\linewidth]{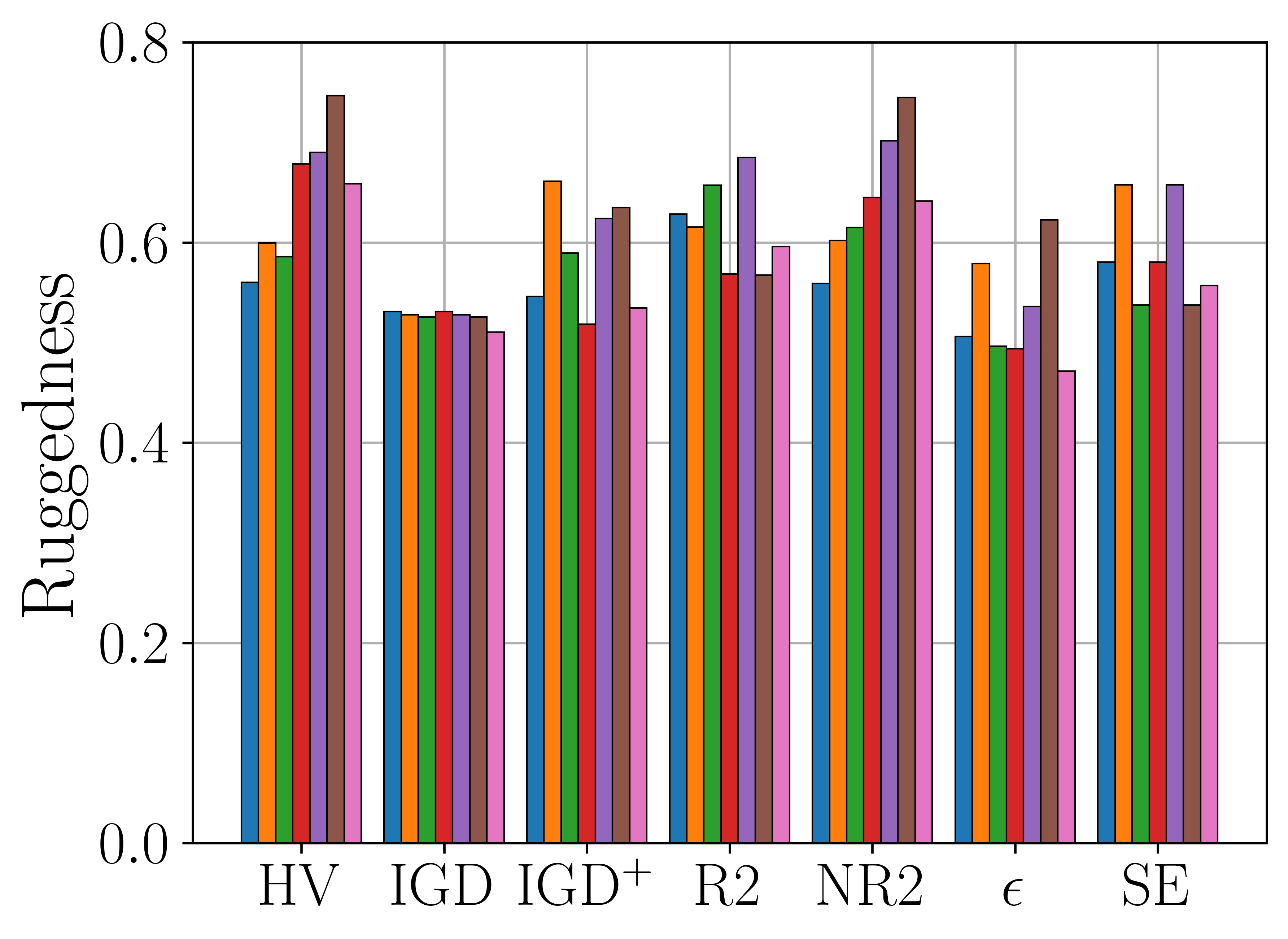}
    \caption{Degree of ruggedness. Each bar indicates the correlation between the quality indicator values of neighbor solutions.}
    \label{fig:rug}
\end{figure}


\pref{fig:rug} shows the ruggedness of each ISSP instance.
As shown in \pref{fig:rug}, the correlation coefficient is high and around $0.5$ even in the minimum case.
This means that the landscape of each ISSP instance is not rugged and is relatively smooth.

As seen from \pref{fig:rug}, the ruggedness is slightly influenced by the type of quality indicator and the shape of the PF.
For example, the HV-SSP with the conventional PFs have lower ruggedness than that with their inverted versions.
Compared to the other ISSP instances, the ruggedness of the IGD-SSP is low and does not significantly depend on the shape of the PF.


\begin{tcolorbox}[sharpish corners, top=2pt, bottom=2pt, left=4pt, right=4pt, boxrule=0.0pt, colback=black!5!white,leftrule=0.75mm,]
\textbf{Main finding}:
The landscape of the ISSP is likely to be relatively smooth in most cases.
\end{tcolorbox}

\subsection{Neutrality}
\label{subsec:neutrality}


\begin{figure}[t]
    \centering
    \includegraphics[width=.87\linewidth]{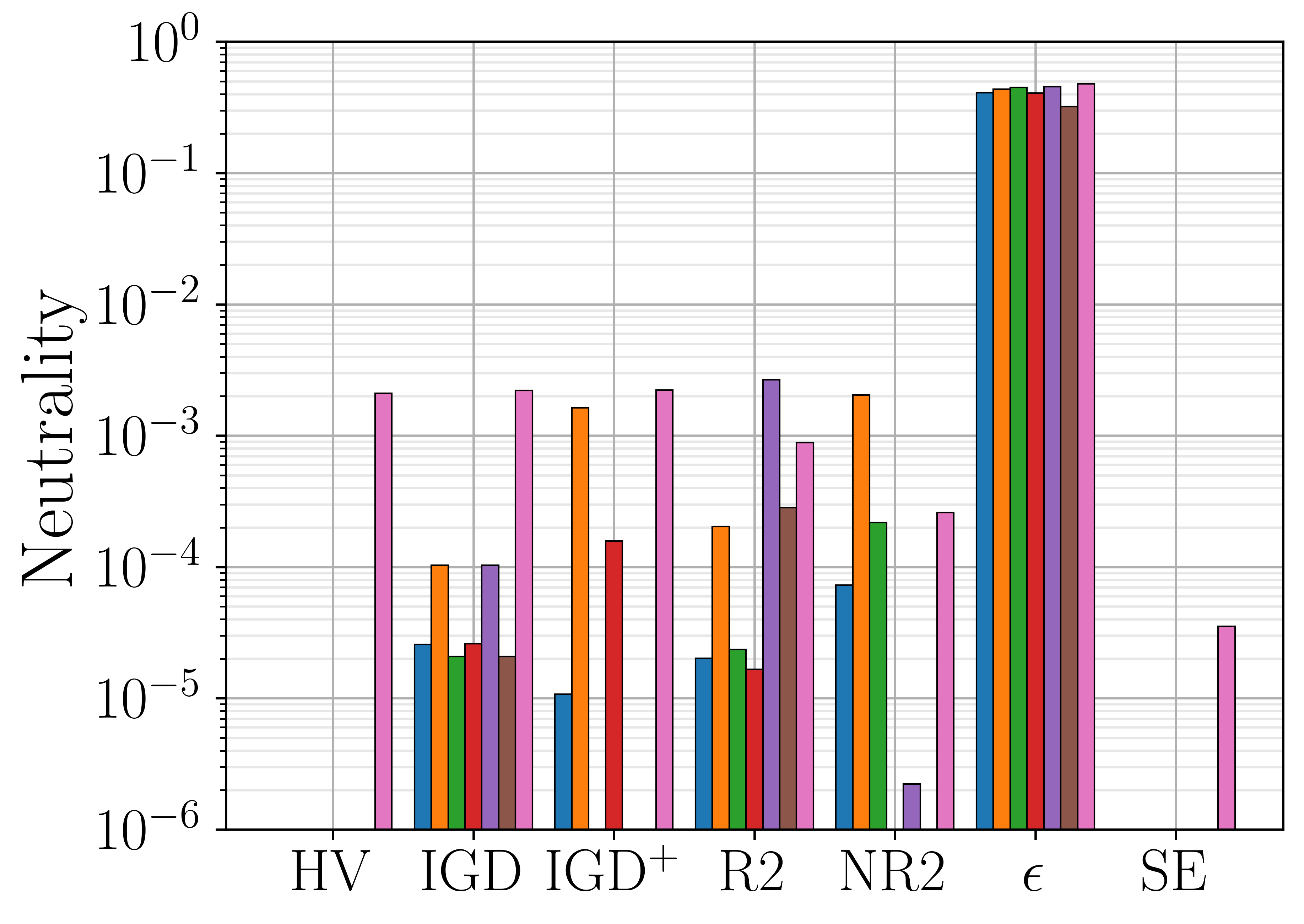}
    \caption{Degree of neutrality. Each bar indicates the average proportion of the number of solutions with the same quality in the neighborhood of each solution.}
    \label{fig:neu}
\end{figure}

\pref{fig:neu} shows the neutral degree of each ISSP instance.
Our results in \pref{fig:neu} indicate that the type of quality indicator and the shape of the PF influence the degree of neutrality.
Neutrality can be found in the results for the IGD-SSP, R2-SSP, and $\epsilon$-SSP independently of the shape of the PF.
High neutrality is observed in the results for the $\epsilon$-SSP with all PFs.
In contrast, the neutral degree is zero in the results for the HV-SSP, IGD$^+$-SSP, NR2-SSP, and SE-SSP in some cases.
Especially, neutrality cannot be found in the results for the HV-SSP and SE-SSP, except for the case of the discontinuous PF.

High neutrality in the results for the discontinuous PF is due to the axisymmetric distribution of $P$ rather than the discontinuity.
Since the method \cite{TianXZCJ18} generates $P$ by transforming perfectly uniform points in $[0, 1]^{d-1}$, the distribution of $P$ is axisymmetric.
Swapping two axisymmetric points also does not influence the quality indicator value of the subset.
This gives neutrality in the landscape of the ISSP.
%

\begin{tcolorbox}[sharpish corners, top=2pt, bottom=2pt, left=4pt, right=4pt, boxrule=0.0pt, colback=black!5!white,leftrule=0.75mm,]
\textbf{Main finding}:
The neutral degree mainly depends on the type of quality indicator.
While the neutral degree is high in the results for the $\epsilon$-SSP, it is zero in the results for the HV-SSP and SE-SSP in most cases.

\end{tcolorbox}

\subsection{Fitness distance correlation (FDC)}
\label{subsec:fdc}


\begin{figure}[t]
    \centering
    \subfloat[Hamming distance]{
    \includegraphics[width=.8\linewidth]{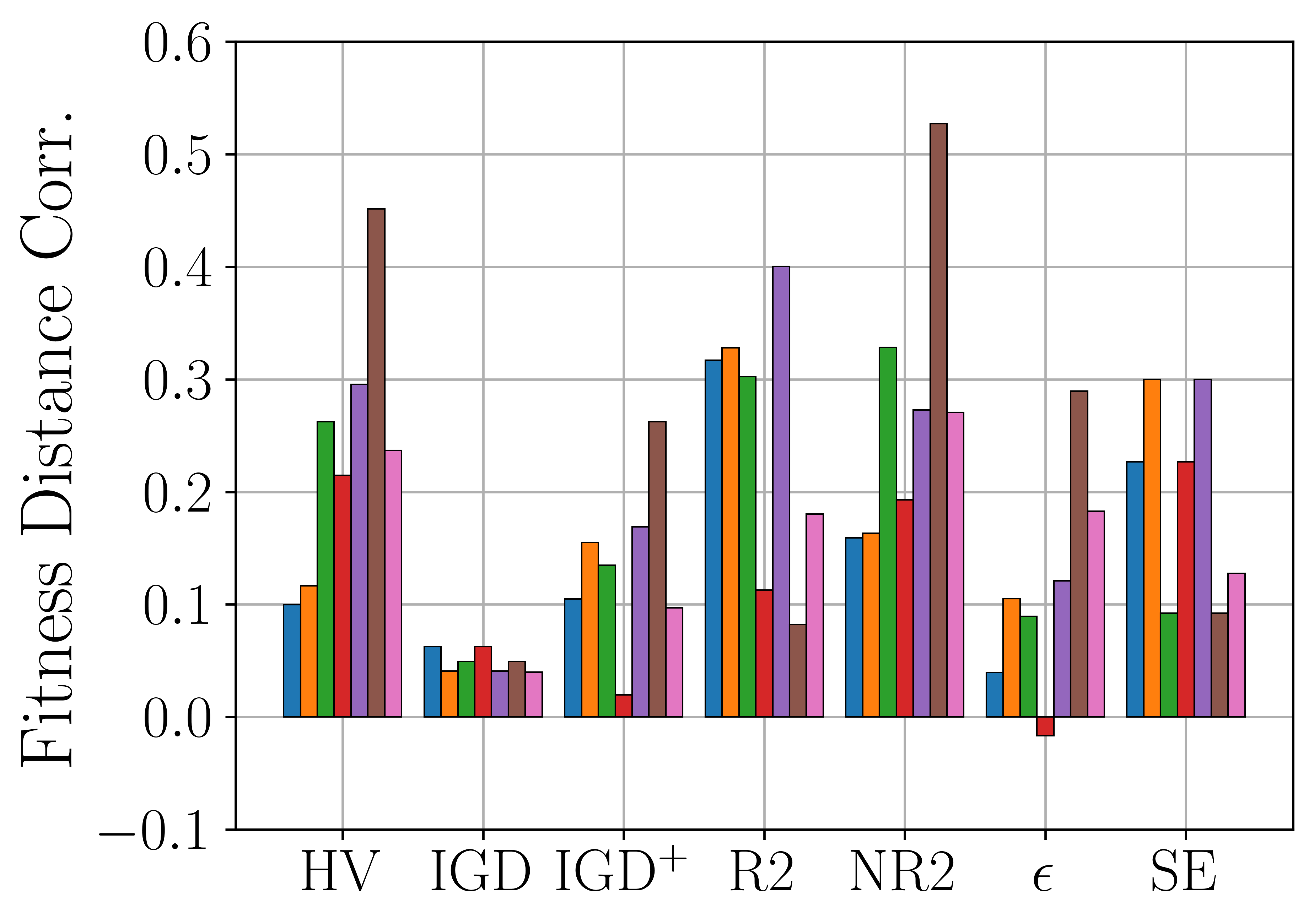}
    }\\
    \subfloat[Wasserstein distance]{
    \includegraphics[width=.8\linewidth]{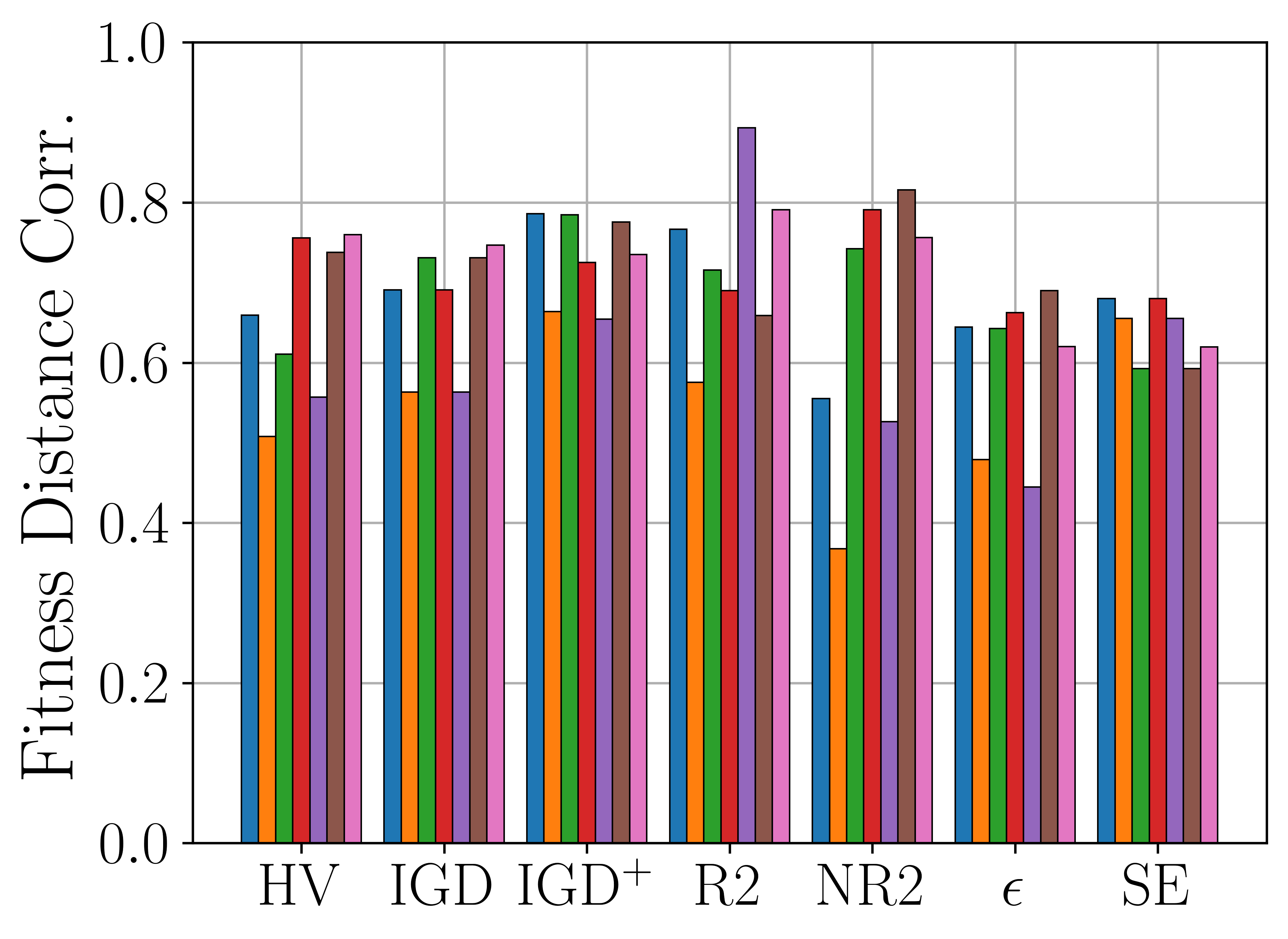}
    }
    \caption{FDC with the two distance measures.}
    \label{fig:fdc}
\end{figure}



Figures \ref{fig:fdc}(a) and (b) show the FDC values based on the Hamming and Wasserstein distance measures, respectively.

\noindent \textit{FDC with the Hamming distance.}
As shown in \pref{fig:fdc}(a), variations are found in the FDC value across ISSP instances.
However, small FDC values are observed in most cases.
Especially, the results for the IGD-SSP exhibit small FDC values for all PFs.
In other words, effective search directions are unlikely to be available in the landscape of the IGD-SSP due to its weak global structure. 


\noindent \textit{FDC with the Wasserstein distance.}
By comparing Figures \ref{fig:fdc}(a) and (b), it is evident that the FDC value measured by the Wasserstein distance is larger than that by the Hamming distance.
In other words, the global structure of the landscape is strong when measuring the distance between two solutions in the objective space $V$ (i.e., the phenotype space) instead of in the binary space $\{0,1\}^n$ (i.e., the genotype space).
This observation supports the validity of the candidate list strategy \cite{KorogiT25}, which restricts local search to swap only two points close to each other in the objective space.


\begin{tcolorbox}[sharpish corners, top=2pt, bottom=2pt, left=4pt, right=4pt, boxrule=0.0pt, colback=black!5!white,leftrule=0.75mm,]
\textbf{Main finding}:
The choice of distance function has a significant impact on the global structure of the landscape of the ISSP.
Finding a better subset could be easy when exploiting the neighborhood relation in the objective space.

\end{tcolorbox}

\subsection{Local optima networks (LONs)}
\label{subsec:lons}

\begin{figure}[t]
    \captionsetup[subfloat]{farskip=2pt,captionskip=1pt}
    \centering
    \newcommand{\widthvar}{0.5}
    \subfloat[HV and linear PF]{\includegraphics[width=\widthvar\linewidth]{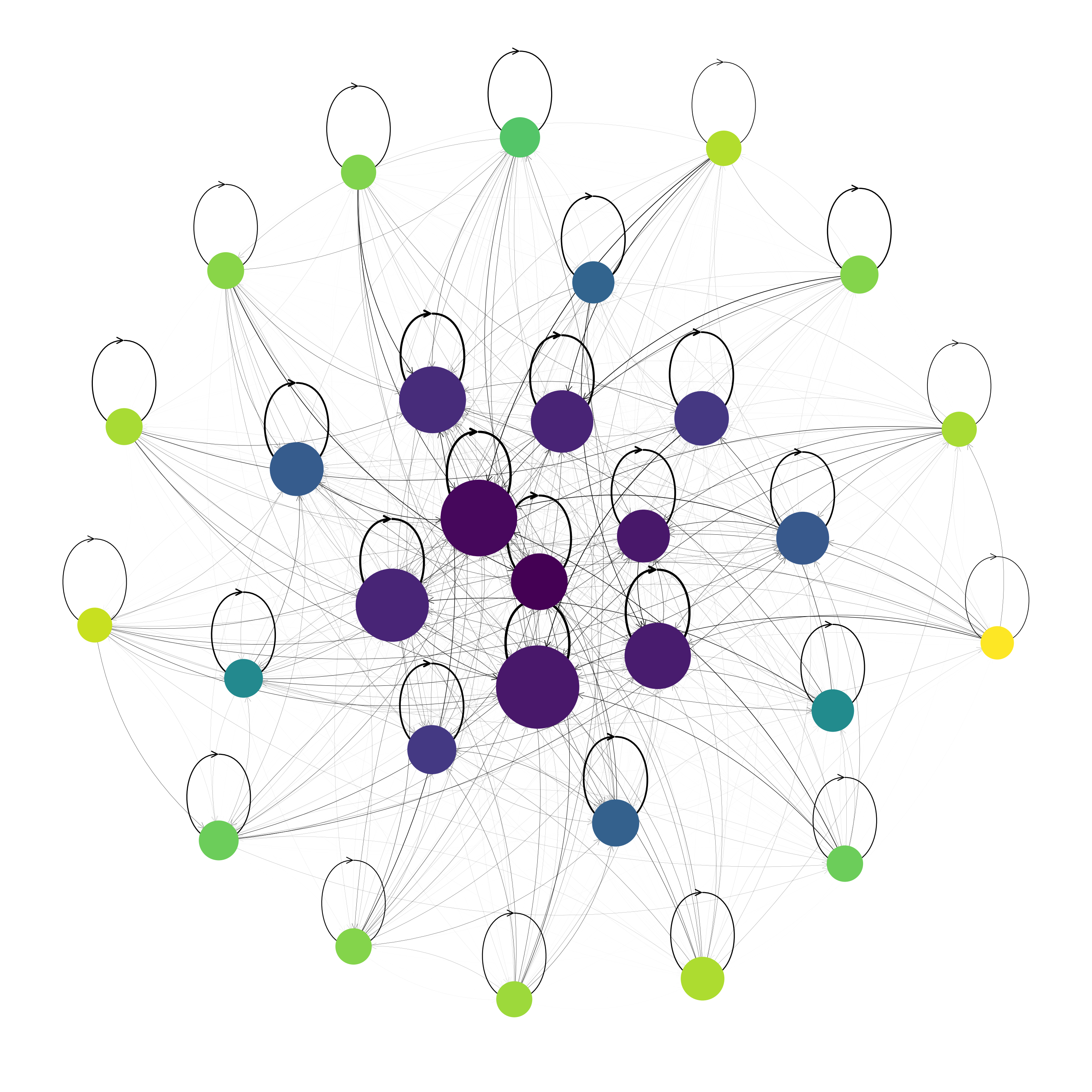}}
    \subfloat[IGD$^+$ and inv-nonconvex PF]{\includegraphics[width=\widthvar\linewidth]{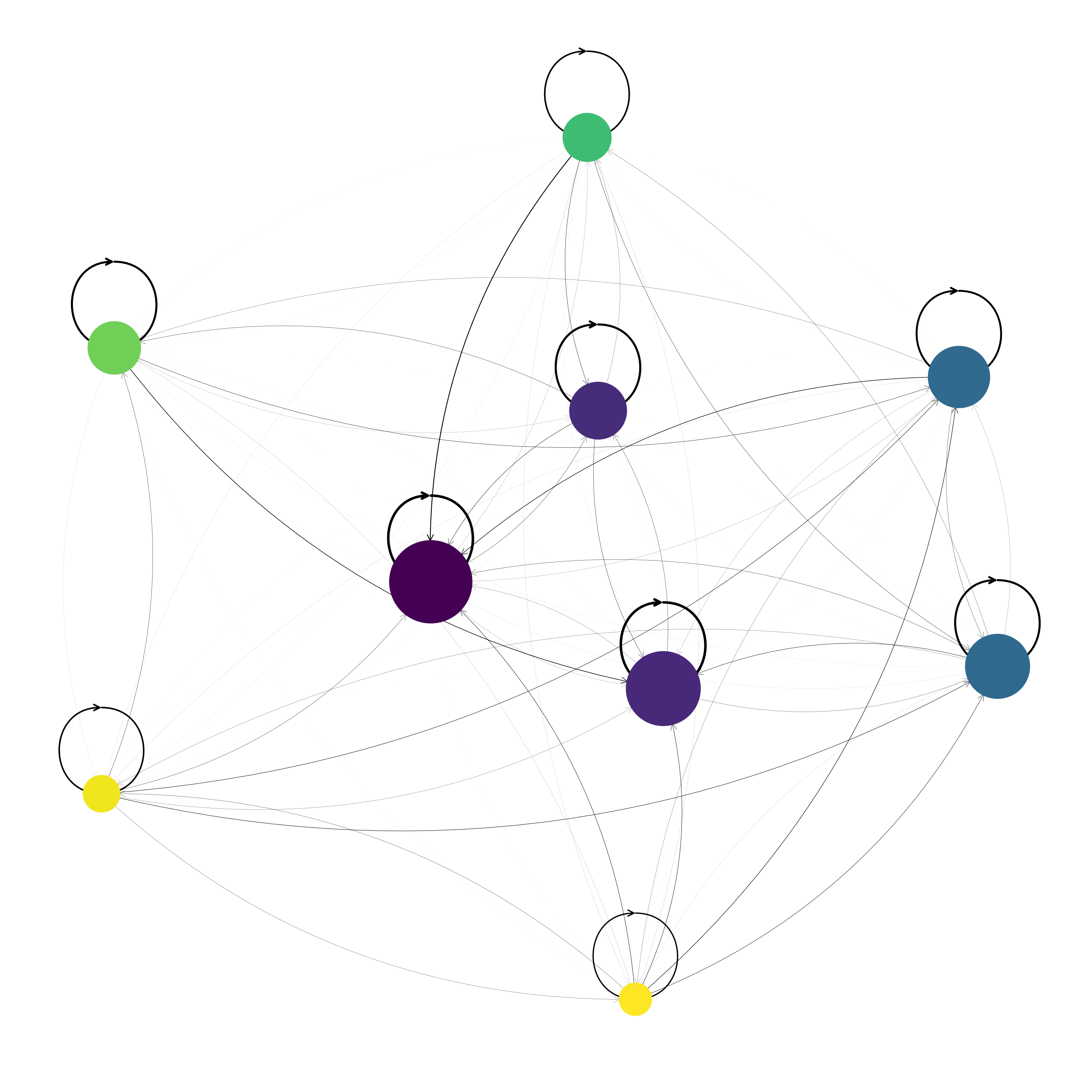}} \\
    \subfloat[SE and convex PF]{\includegraphics[width=\widthvar\linewidth]{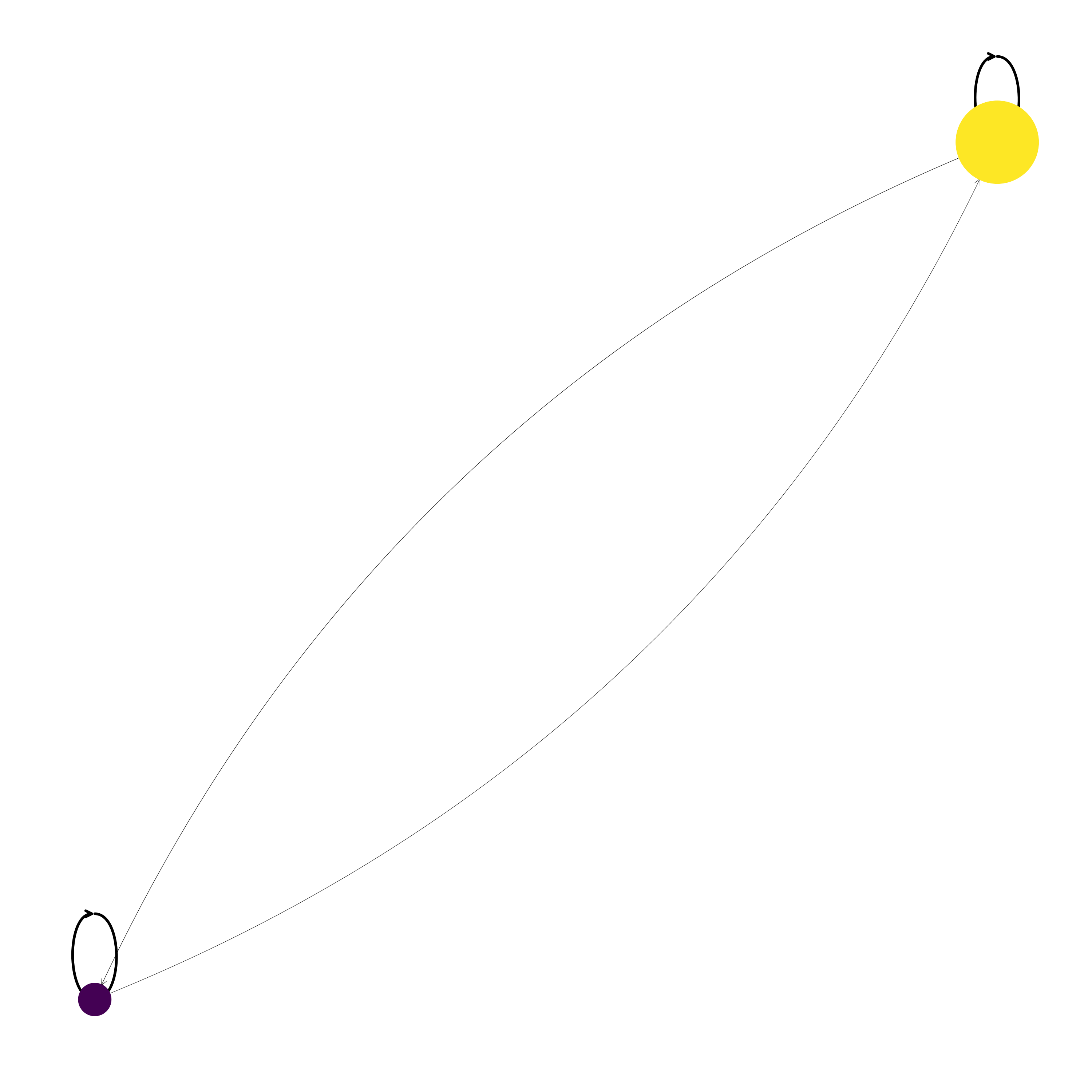}}
    \subfloat[$\epsilon$ and convex PF]{\includegraphics[width=\widthvar\linewidth]{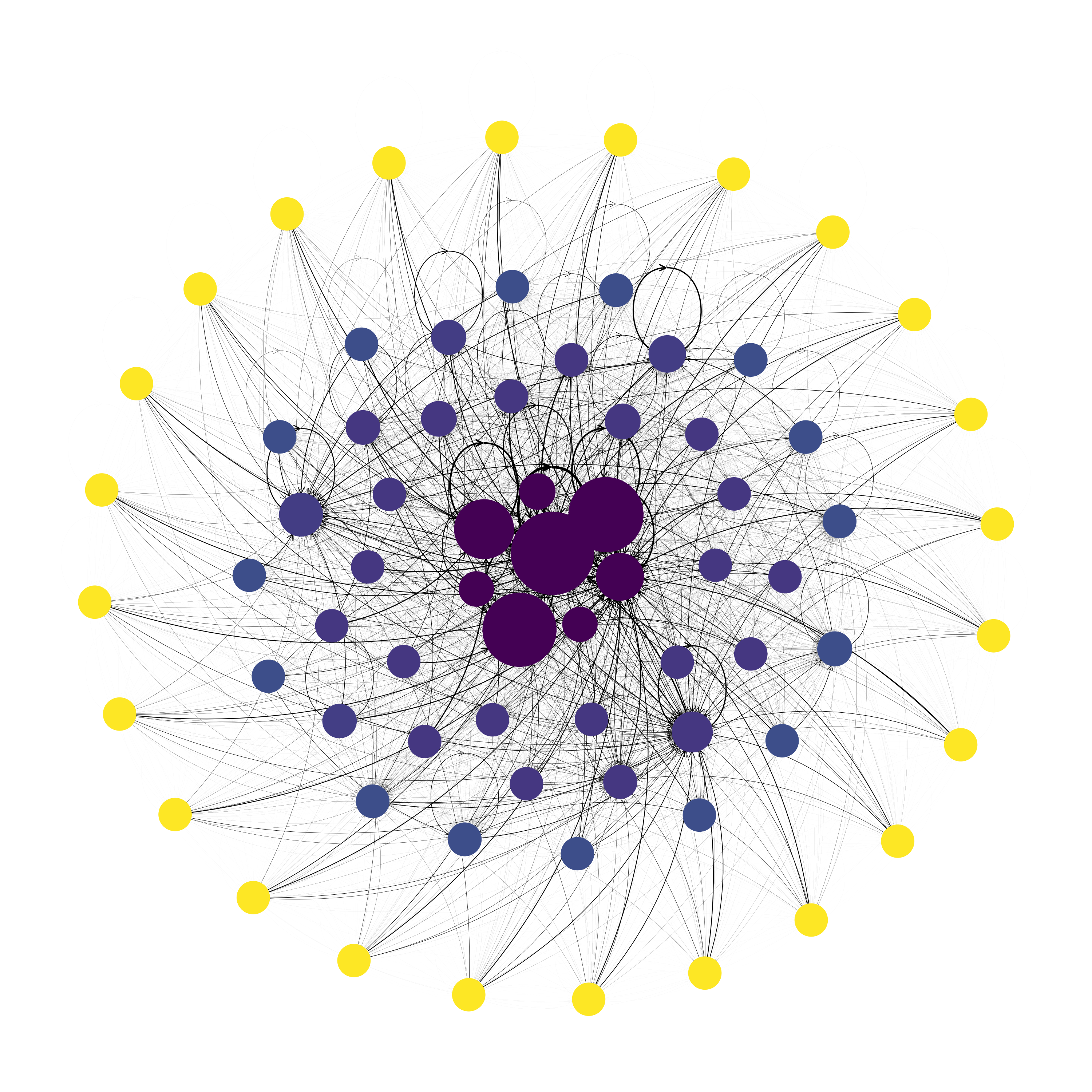}}

    \caption{Examples of LONs. Each local optimum corresponds to a node. The colors and sizes of the nodes and the width of the escape edges visualize the structure of the problem.}
    \label{fig:lo-net}
\end{figure}

Since a LON can be computed for each ISSP instance, 49 LONs are available for the 49 ISSP instances.
Due to the paper length limitation, \pref{fig:lo-net} shows only four notable LONs for four ISSP instances, where Figures \ref{fig:lo-net}(a)--(d) show the results for the HV-SSP with the linear PF, IGD$^+$-SSP with the inv-nonconvex PF, SE-SSP with the convex PF, and $\epsilon$-SSP with the convex PF, respectively.
In \pref{fig:lo-net}, the color of each node represents a normalized quality indicator value by the maximum and minimum ones in all local optima.
A darker color indicates that the corresponding node is of better quality.
The width of each edge shows weight, which represents the transition probability.
%

As shown in \pref{fig:lo-net}, the distribution of local optima and the size of basins of attraction significantly differ depending on the type of quality indicator.
As seen from \pref{fig:lo-net}(a), the HV-SSP  with the linear PF has a typical ``big-valley'' structure~\cite{BoeseKM94}, where poor local optima surrounds the single global optimum.
The LON in \pref{fig:lo-net}(b) is sparser than that in \pref{fig:lo-net}(a).
We do not describe the results in detail, but LONs like \pref{fig:lo-net}(b) are found for the IGD-SSP, R2-SSP, and NR2-SSP.
Since the SE-SSP with the convex PF has only two local optima as shown in \pref{fig:num-optima}(b), the LON in \pref{fig:lo-net}(c) consists of only two nodes.
Sparse LONs like \pref{fig:lo-net}(c) are observed for ISSP instances that have only a few local optima.
We found that the $\epsilon$-SSP has unique LONs like \pref{fig:lo-net}(d).
In \pref{fig:lo-net}(d), an edge exists from poor outer nodes to better center nodes.


\begin{tcolorbox}[sharpish corners, top=2pt, bottom=2pt, left=4pt, right=4pt, boxrule=0.0pt, colback=black!5!white,leftrule=0.75mm,]
\textbf{Main finding}:
The type of quality indicator significantly influences the structure of LONs.
For example, while LONs for the HV-SSP appear simple, LONs for the $\epsilon$-SSP are complicated.

\end{tcolorbox}






\begin{figure*}[t]
    \centering
    \subfloat[GS-F]{
    \includegraphics[width=.325\linewidth]{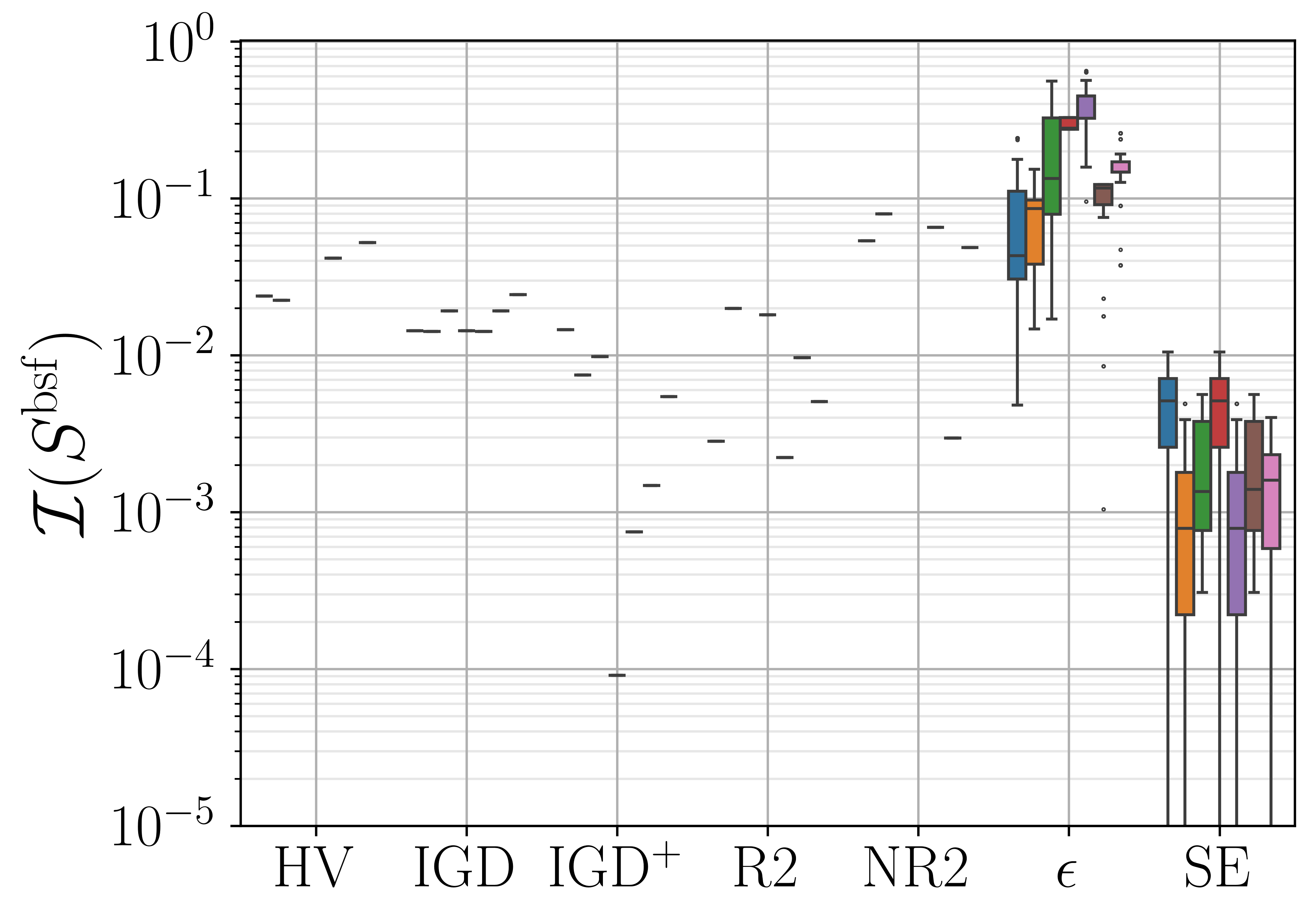}
    }
    \subfloat[GS-B]{
    \includegraphics[width=.325\linewidth]{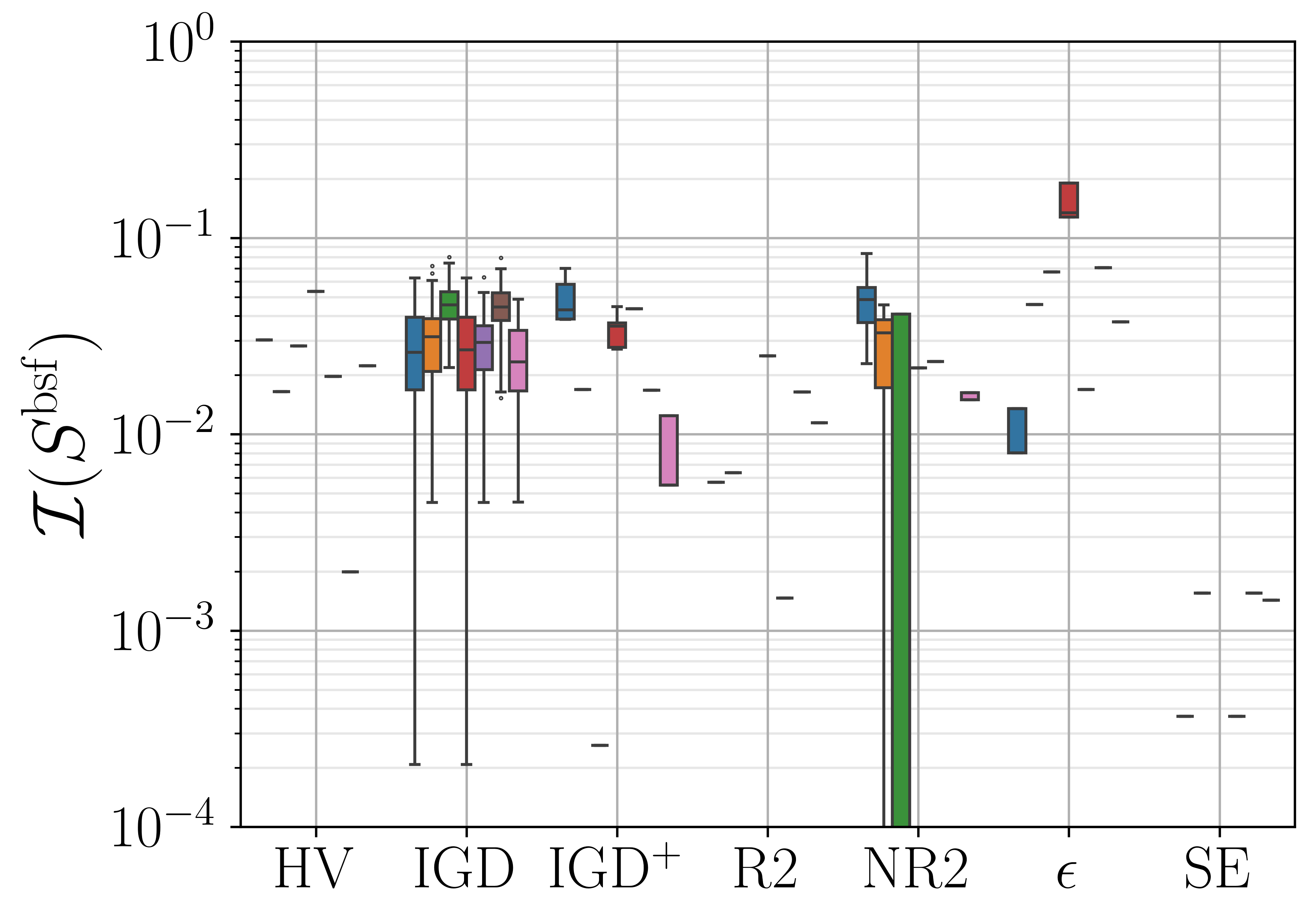}
    }
    \subfloat[LS]{
    \includegraphics[width=.325\linewidth]{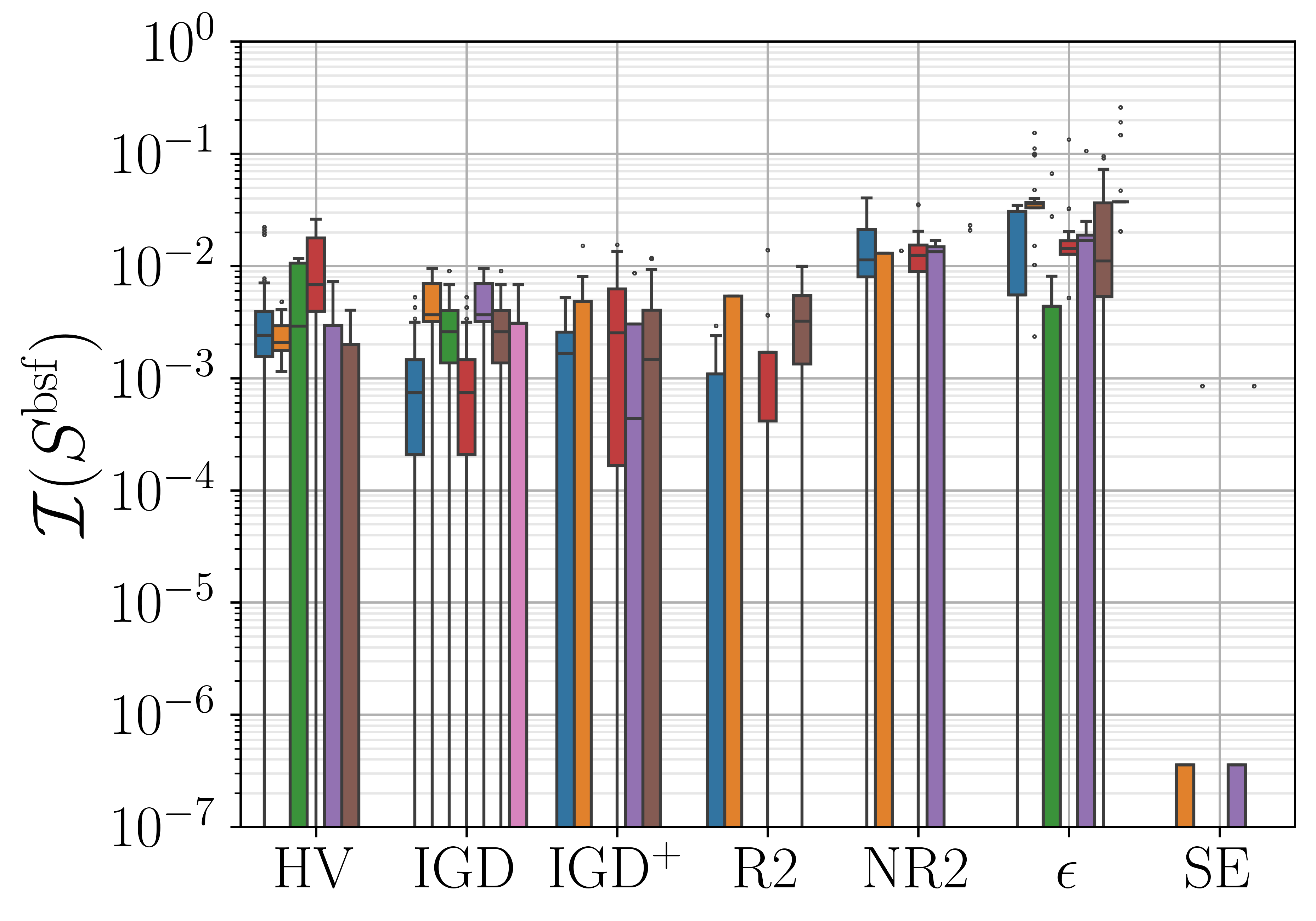}
    }
    \caption{Performance of GS-F, GS-B, and LS on the 49 ISSP instances. The best-so-far quality indicator values $\mathcal{I}(S^{\mathrm{bsf}})$ are min-max normalized based on the values across all subsets.}
    \label{fig:alg}
\end{figure*}

\subsection{Performance of subset selection methods}
\label{subsec:alg}

Unlike the other sections, this section discusses how the performance of subset selection methods relates to the landscape of the ISSP.
We applied the following three basic subset selection methods to the 49 ISSP instances: forward greedy search (GS-F), backward greedy search (GS-B), and best-improvement local search (LS).
Starting from $S = \emptyset$, GS-F repeatedly adds the point to $S$ that improves the quality of $S$ the most until $|S| = k$.
In contrast to GS-F, starting from $S = P$, GS-B repeatedly removes the worst point from $S$ in terms of the contribution to the quality of $S$ until $|S| = k$.
For LS, see Section \ref{subsec:prelim:land-analysis}.
For subsets with the same quality, ties are broken by random selection.


\pref{fig:alg}(a)--(c) show the performance of GS-F, GS-B, and LS on the ISSP instances, respectively.
For each ISSP instance, \pref{fig:alg} shows the distribution of the 101 best-so-far quality indicator values $\mathcal{I}(S^\mathrm{bsf})$ found by each method over 101 runs, where the quality indicator values are normalized based on the maximum and minimum values for all subsets.

\noindent \textit{Results of GS-F and GS-B.}
Variations in the performance of GS-F and GS-B are found in \pref{fig:alg}(a) and (b). 
Those are due to the existence of neutrality in the landscape of the ISSP, as demonstrated in \pref{subsec:neutrality}.
Only one exception is the results of GS-F on the SE-SSP, where variations in its performance are because $\mathrm{SE}(\{p\}) = 0$ for any $p \in P$ at the beginning of the search (i.e., $S=\emptyset$).
GS-F and GS-B perform poorly on the $\epsilon$-SSP, and they can find only subsets far from global optima in terms of quality.



\noindent \textit{Results of LS.}
The results in \pref{fig:alg}(c) indicate that the SE-SSP is the easiest to solve for LS. 
This is mainly because the SE-SSP has the fewest number of local optima and weak neutrality in most cases, as observed in Sections \ref{subsec:num-optima} and \ref{subsec:neutrality}, respectively.
Overall, the distribution of the best-so-far quality indicator values obtained by LS is related to on the number of local optima.

As demonstrated in \pref{subsec:corr-sol-rank}, high correlations are found between the HV-SSP and NR2-SSP.
However, based on the results of LS in \pref{fig:alg}(c), this does not mean that the difficulty in finding a global optimum is equivalent for the HV-SSP and NR2-SSP, especially when considering the convex and nonconvex PFs.
Similarly, the $\epsilon$-SSP is harder to solve for LS than the IGD$^+$-SSP despite their high correlation.
This observation provides a clue to design an efficient subset selection method for the $\epsilon$-SSP.
For example, a better subset in the $\epsilon$-SSP may be found by LS that uses the best subset in the IGD$^+$-SSP as the initial subset.




\begin{tcolorbox}[sharpish corners, top=2pt, bottom=2pt, left=4pt, right=4pt, boxrule=0.0pt, colback=black!5!white,leftrule=0.75mm,]
\textbf{Main finding}: The number of local optima and neutral degree could have a large impact on the performance of the subset selection methods.
The degree of correlation between two ISSP instances is unlikely to explain how similarly LS performs on them.





\end{tcolorbox}




\subsection{Influence of the number of objectives $d$}
\label{subsec:d}

Throughout this paper, we have analyzed the landscape of the ISSP with $d=3$. 
This section investigates the influence of $d$ on the landscape of the ISSP. 
In this section, $d$ is set to $2, 3, ..., 6$.
We set the subset size $k$ to $7$ such that $k > d$ for any $d \in \{2, 3, ..., 6\}$.
If $k \leq d$, an optimal subset is obvious, where it should include only extreme points on the PF.
The point set size $n$ was set to $30$ such that all subsets can be fully enumerated, where the size of the solution space is $\binom{n}{k} = 2\,035\,800$ in this case. 
We fixed $n$ and $k$ for any $d$ to investigate only the influence of $d$.
Since $n$ cannot be fixed to about $30$ for the discontinuous PF, we remove it from our analysis.

Figures \ref{supfig:d:distr}--\ref{supfig:d:alg} in the supplementary file show the distribution of quality indicator values (Figure \ref{supfig:d:distr}), correlation coefficient between the rankings of all subsets for each pair of quality indicators (Figure \ref{supfig:d:corr}), number of global and local optima (Figure \ref{supfig:d:num-optima}), degree of ruggedness (Figure \ref{supfig:d:rug}), degree of neutrality (Figure \ref{supfig:d:neu}), FDC value (Figure \ref{supfig:d:fdc}), performance of the three subset selection methods (Figure \ref{supfig:d:alg}), respectively when $d \in \{2, 3, \dots, 6\}$.
Figures \ref{supfig:d:distr}--\ref{supfig:d:alg} correspond to Figures \ref{fig:distr-qi}, \ref{fig:corr-sol-rank}, \ref{fig:num-optima}, \ref{fig:rug}, \ref{fig:neu}, \ref{fig:fdc}, and \ref{fig:alg}, respectively.


We do not explain Figures \ref{supfig:d:distr}--\ref{supfig:d:alg} in detail due to the paper length
limitation, but they indicate that the setting of $d$ slightly influences the landscape of the ISSP.
For example, as shown in Figure \ref{supfig:d:num-optima}, the number of global and local optima increases for the $\epsilon$-SSP as $d$ increases.
As seen from Figure \ref{supfig:d:neu}, except for the results for the IGD-SSP, high neutrality is observed for a large $d$.
However, overall, the conclusions of Sections \ref{subsec:ruggedness}--\ref{subsec:alg} are not significantly influenced by the setting of $d$.

%


\begin{tcolorbox}[sharpish corners, top=2pt, bottom=2pt, left=4pt, right=4pt, boxrule=0.0pt, colback=black!5!white,leftrule=0.75mm,]
\textbf{Main finding}: 
The setting of $d$ does not have a significant impact on the landscape of the ISSP. 
\end{tcolorbox}

\section{Conclusion}
\label{sec:conclusion}

We have analyzed the landscape of the 49 ISSP instances with the 7 quality indicators and 7 PFs by means of traditional landscape measures and LONs.
In summary, we found that the landscape of the ISSP significantly depends on the type of the quality indicators and shape of the PF.
For example, our results show that the $\epsilon$-SSP has many global and local optima, whereas other ISSP instances have only one local optimum for some PFs.
High neutrality is observed in the results for the $\epsilon$-SSP, whereas no neutrality is found in the results for the HV-SSP and SE-SSP in most cases.
We also demonstrated that three basic subset selection methods (GS-F, GS-B, and LS) struggle on the $\epsilon$-SSP due to these unique landscape properties.
%





We believe that our findings contribute to the design of efficient subset selection methods for the ISSP.
%
Based on the results of FDC with the Hamming and Wasserstein distance measures, it is promising to exploit neighborhood relations in the objective space as in \cite{KorogiT25}.
Our analysis was based on the small point set size $n$ and subset size $k$ so that we can fully enumerate all subsets.
Landscape analysis of the ISSP with large $n$ and $k$ values by using approximate approaches is needed in future work.
It is also important to investigate the influence of $k$ on the landscape of the ISSP.




\begin{acks}
   This work was supported by JSPS KAKENHI Grant Number \seqsplit{25K03194} and \seqsplit{23H00491}.
\end{acks}

\bibliographystyle{ACM-Reference-Format}
\bibliography{reference} 










\end{document}